\documentclass[letterpaper, 10 pt, conference]{ieeeconf}  

\IEEEoverridecommandlockouts                              

\usepackage{graphics} 
\usepackage{epsfig} 
\usepackage{graphicx}
\usepackage{times} 
\usepackage{amsmath} 
\usepackage{amssymb}  
\usepackage{caption}
\usepackage{balance}
\usepackage{subfigure}
\usepackage{cite}

\title{Simultaneous Localization, Mapping, and Manipulation\\
for Unsupervised Object Discovery}

\author{Lu Ma, Mahsa Ghafarianzadeh, Dave Coleman, Nikolaus Correll, and Gabe Sibley*
\thanks{*This work was supported by Google, Toyota and NASA.}
\thanks{Computer Science Department, University of Colorado at Boulder, CO, USA {\tt\small lu.ma|mahsa.ghafarianzadeh|gsibley|
\tt\small david.t.coleman|ncorrell@colorado.edu}}
}

\begin{document}

\maketitle
\thispagestyle{empty}
\pagestyle{empty}

\begin{abstract}
We present an unsupervised framework for simultaneous appearance-based object discovery, detection, tracking and reconstruction using RGBD cameras and a robot manipulator.  The system performs dense 3D simultaneous localization and mapping concurrently with unsupervised object discovery. Putative objects that are spatially and visually coherent are manipulated by the robot to gain additional motion-cues.  The robot uses appearance alone, followed by structure and motion cues, to jointly discover, verify, learn and improve models of objects.  Induced motion segmentation reinforces learned models which are represented implicitly as 2D and 3D level sets to capture both shape and appearance.  We compare three different approaches for appearance-based object discovery and find that a novel form of spatio-temporal super-pixels gives the highest quality candidate object models in terms of precision and recall. Live experiments with a Baxter robot demonstrate a holistic pipeline capable of automatic discovery, verification, detection, tracking and reconstruction of unknown objects.
\end{abstract}

\section{INTRODUCTION}
Automatic object model acquisition is a fundamental task in robotics.  Recent state-of-the-art object detection and recognition systems are based on pre-trained detectors \cite{bo2013unsupervised,endres2009unsupervised,herbst2011rgb,meger2008curious} and features \cite{morisset2009leaving,lai2011large}. For object learning, most techniques reconstruct sparse 2D models \cite{herbst2011toward,collet2013exploiting} together with dense 3D shape models. The focus of this paper is an unsupervised pipeline for discovering unknown objects in a scene without any prior information and using robots with manipulators to  verify  and reinforce model acquisition by robot-object interaction (e.g. poking, grasping). Hence, a pre-trained object detection technique is not suitable here -- indeed, the proposed method could be used to train such systems.

2D object appearance is highly informative for image based spatio-temporal segmentation \cite{grundmann2010efficient,ghafarianzadeh2014unsupervised}. Typically, per-frame image segmentation results are combined with temporal information via 2D segment tracking.  Instead, we find that using appearance and dense spatio-temporal cues produces better image segmentation results, which in turn yield a high quality set of likely candidate objects. However, these cues are not sufficient for the verification of object-hood, especially for static objects \cite{ma2014unsupervised,mason2012object}.   In the proposed approach, a mobile robot relies on dense 3D SLAM to manipulate putative objects, thereby adding possible motion cues to verify object-hood.

\begin{figure}
\centering
\includegraphics[keepaspectratio=true,scale=0.30]
{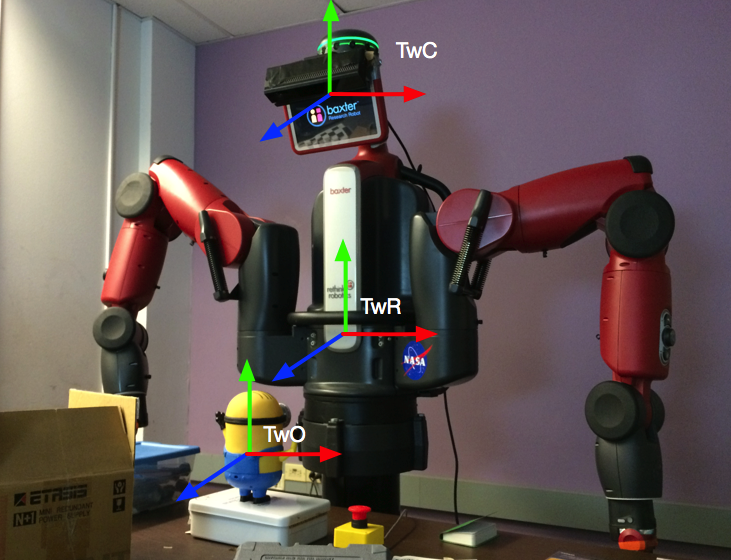}
\caption{The dual-arm Baxter robot equipped with a RGB-D camera automatically discovers and learns dense 3D model of unknown objects in the scene by manipulation.}
\label{fig:Platform}
\end{figure}

Once an object has been verified by poking or grasping, a dense detection, tracking and reconstruction technique is applied to learn 2D and 3D implicit models of appearance and shape.  In subsequent steps, learned models are automatically detected, tracked, and refined.  

Related work demonstrates dense tracking and reconstruction to generate 3D models of static or moving objects \cite{renstar3d,dame2013dense,bao2013dense}. However, these techniques require a bounding box (or a pre-trained object detector) as system input to define the target. 

This paper extends prior work on {\em unsupervised object discovery, detection, tracking and reconstruction} (DDTR)\cite{ma2014unsupervised} with a novel form of appearance based object discovery and also uses a robot manipulator to verify candidate objects.  A complete dense 3D model is generated once the robot has full observation of an object.

The proposed framework is different from most existing frameworks as: 1) it does not require prior information for object discovery~\cite{meger2008curious,bo2013unsupervised}; 2) it builds models and also subsequently detects, tracks and refines {\em dense} 2D and 3D models\cite{bo2013unsupervised,triebel2010segmentation,endres2009unsupervised,herbst2011rgb,meger2008curious,modayil2004bootstrap} \cite{collet2013exploiting};  3) it learns high quality 3D models of unknown objects via robot-object interaction; 4) it is also a simultaneous localization and mapping system (SLAM), which enables the potential of autonomous object discovery in a large environment; 5) the final learned model of an object can be used for future object detection tasks, and shared with other robots. 

The remainder of this paper is structured as follows: Section 2  briefly covers preliminaries and introduces our approach. In the following sections we cover appearance-based object discovery, motion-based objects verification, tracking, reconstruction and system integration. Section 6 tests and discusses the system performance. Section 7 addresses failure cases and future work and Section 8 draws conclusions.

\section{Overview}
\subsection{Preliminaries}
The proposed system (Fig.\ref{fig:Platform}) uses a dual-arm Baxter robot  with a calibrated RGBD camera $\mathcal{C}$ mounted on its head, where the world pose of the center of the robot is $T_{wR}\in SE(3)$. The camera $\mathcal{C}$ is calibrated with respect to the robot center, with extrinsic matrix $T_{wC} \in SE(3)$, and intrinsic matrix $K_{i}$. The transformation between $T_{wR}$ and $T_{wC}$ is calibrated and considered as a known parameter. An object in the scene is noted as $O_{i}$, where its world pose $T_{wO}$ is noted as the center point of the object, and can be computed by $T_{wO} = T_{wC} \oplus T_{CO}$.

A 3D point in the camera frame is denoted as $x_{c}=(x,y,z)^{\top}$ and projects onto the image plane as  $\pi(x_{c})=(x/z,y/z)^{\top}$ (i.e., dehomogenization). A 2D point $\mathbf{u}=(u,v)^{\top}$ in the image plane can be back-projected to a 3D point by $\chi= \frac{1}{d} \cdot K_{i}^{-1}\cdot (\mathbf{u};1)$ with a depth value $d$. The system uses $T_{wr}^i$ and $T_{wl}^i$ to represent the camera pose in the reference frame (time $k-1$) and the live frame (time $k$). 

\begin{figure}
\centering
\includegraphics[keepaspectratio=true,scale=0.18]
{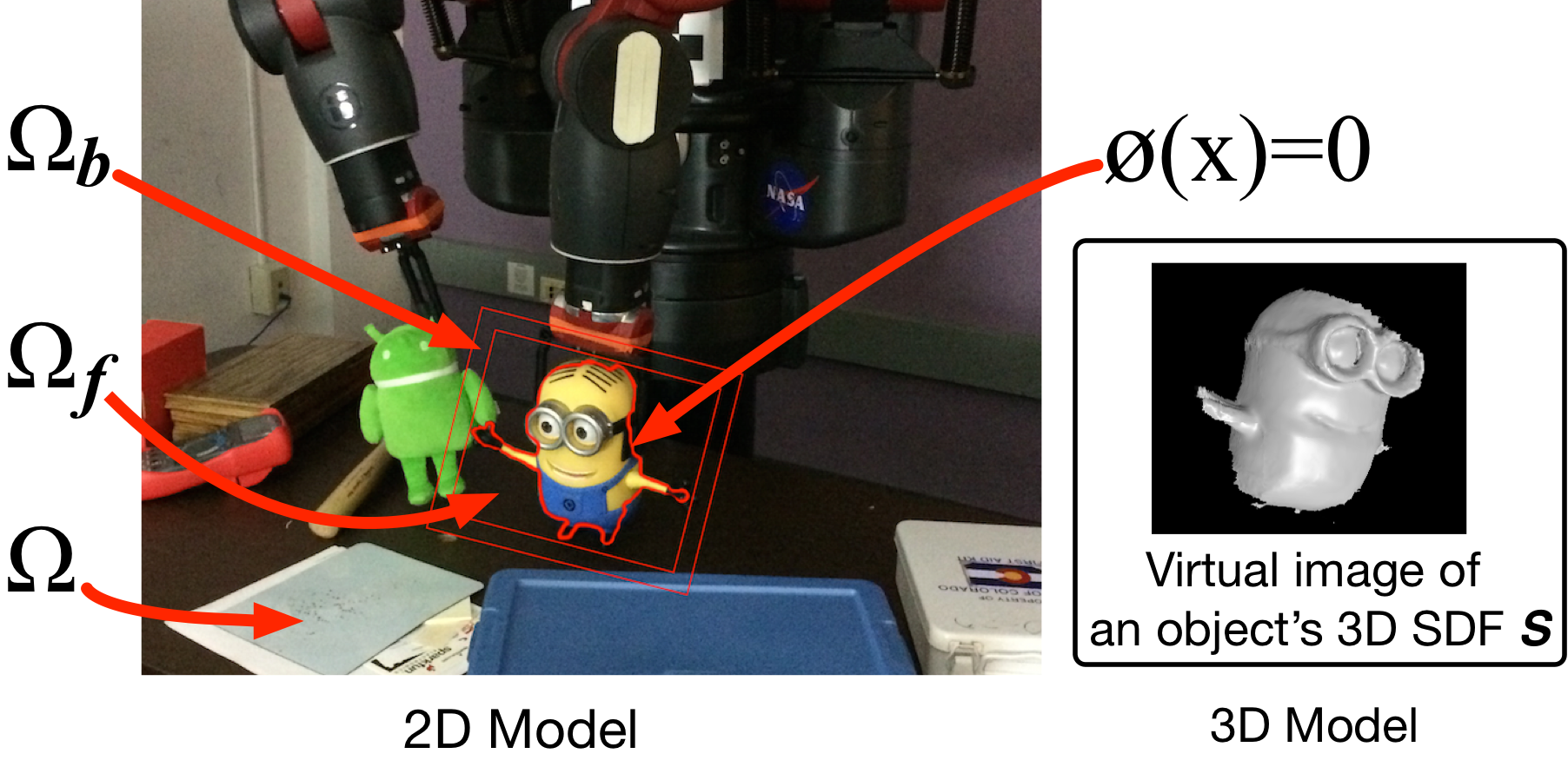}
\caption{An Object is represented implicitly with a 2D and a 3D model. The 2D model is represented with a RGB histogram and a 2D SDF  $\Phi(x)$. The zero level set of $\Phi(x)$ segments the image domain $\Omega$ into a foreground $\Omega_{f}$ and a background $\Omega_{b}$. The 3D model of the object is represented by the 3D SDF and can be rendered via ray casting and phong shading.}
\label{fig:Representation}
\end{figure}

We denote learned objects as $O=O_{\emptyset}, O_{1},O_{2},\cdots,O_{n}$. Each object is represented by both a 2D and a 3D model. Figure \ref{fig:Representation} shows the representation of an object's 2D and 3D model. 

The system uses shape (2D contour) and appearance (RGB histogram) to represent an object's 2D model. A level-set embedding function $\Phi(x_{i})$, namely a 2D signed distance function (SDF), is used to implicitly represent an object's 2D contour (shape). $x_{i}=x_{i1},x_{i2}\cdots,x_{in}$ is the set of pixel locations in the coordinate frame of $O_{i}$. The zero level-set of the object's 2D SDF $\Phi(x)=0$ is used to represent the shape (2D contour). The 2D contour $\Phi(x)=0$ segments the image domain $\Omega$ into a foreground $\Omega_{f}$ and a background $\Omega_{b}$, with appearance models $M_{f}$ and $M_{b}$ respectively. Notice that $\Omega^{\Phi}_{f}$ and $\Omega^{\Phi}_{b}$ is used to denote the foreground (pixels inside $\Phi(x)=0$) and the background segments by the zero level-set of $\Phi(x)$.

The system uses the truncated signed distance function (TSDF) to represent the 3D model $\mathcal{S}_{i}$ of an object $O_{i}$, Where the shape (geometry) of an object can be rendered by ray casting $\mathcal{S}_{i}$ in different views. The 3D SDF $\mathcal{S}$ is stored as a $n\times n\times n$ volume cube, where $n$ is the number of voxels in one dimension. The size of each voxel is $v_{m}=\frac{2r}{n}$, where $r$ is the radius of the volume. The 3D SDF is initialized and updated by SDF fusion. 

Notice that the 2D contour ($\Phi(x)=0$) is actually the boundary of the projection of $\mathcal{S}_{i}$. By using the 2D and 3D models, the system represents objects at different levels and performs 2D\&3D tracking and reconstruction Simultaneously during object learning.

\subsection{System Structure}
\begin{figure}
  \centering \includegraphics[width=0.32\textwidth]{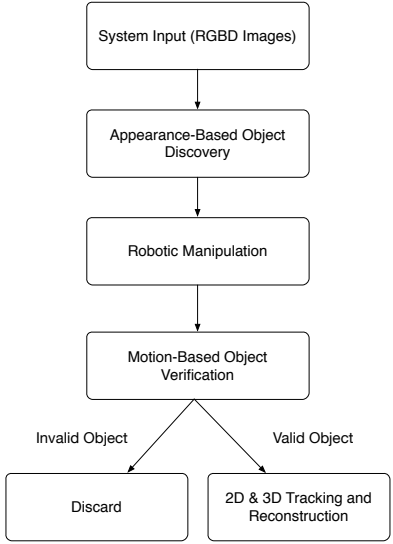}
\caption{System flow chart}
\label{fig:FlowChart}
\end{figure}

The proposed pipeline goes through the following stages (Fig \ref{fig:FlowChart}):

\emph{Initialization}. The system is initialized by creating a 3D model of the first frame (noted as the dominant object $O_{\emptyset}$) from the input RGBD video. We use a $512\times512\times512$ volume for $\mathcal{S}_{i}$. 

\emph{Appearance-based Object Discovery}. For the first 200$-$300 frames, the robot moves its head (the camera) to explore the scene which collects enough information for the appearance-based object discovery approach. While the robot explores the scene, the system tracks and updates $O_{\emptyset}$ frame by frame, and discovers (Section \ref{sec:Appearance-Based-Object-Discovery}) a set of candidate object contour $C^{'}=C^{'}_{1},C^{'}_{2}, \cdots, C^{'}_{n}$ by the appearance-based object discovery approach.

\emph{Robotic Manipulation}. Given a candidate objects contour set $C^{'}$, the system verifies the candidate object $O^{'}_{i}$ (defined by $C^{'}_{i} \in C^{'}$) by applying motion (grasping, poking, etc) to $O^{'}_{i}$. This allow the system generate a verified objects set $O=O_{1},O_{2}\cdots,O_{n}$.

\emph{Motion-based Object Verification}. For a verified object $O_{j}\in O$, the system generates its contour $C_{j}$ by extracting portions of the scene that fail to match with $O_{\emptyset}$ via the ICP+RGB algorithm. The system saves $C_{j}$ to the discovered objects contour set $C$. 

\emph{2D Tracking and 2D Reconstruction}. For a discovered object $O_{j}$ with a contour $C_{j} \in C$, the system matches $C_{j}$ with $O_{i}$ by the appearance, shape and motion cues. This 2D tracking result gives a rough pose estimation of $O_{j}$ in the image domain of $C_{j}$ where $C_{j}$ defines a foreground domain $\Omega_{f}^{j}$. 2D reconstruction is then achieved by updating $\Phi(x_{i})$ to $\Phi^{'}(x_{i})$ in $\Omega_{f}^{j}$ via LSE.  

\emph{3D Tracking and 3D Reconstruction}. The system uses the ICP+RGB pose estimator for 3D tracking, which estimates the relative camera pose between $\Omega_{f}^{\Phi}$ and $\Omega_{f}^{\Phi'}$. Once $O_{i}$ is tracked, the system updates $\mathcal{S}_{i}$ by fusing every pixel from $\Omega_{f}^{\Phi'}$ into $\mathcal{S}_{i}$ via SDF fusion. 

\section{Unsupervised Object Discovery}
The system uses an appearance driven, motion verification pipeline for unsupervised object discovery.

\subsection{Appearance-Based Object Discovery}
\label{sec:Appearance-Based-Object-Discovery}
The system uses \cite{ghafarianzadeh2014unsupervised} to discover candidate objects in the RGB video sequence. This technique generalizes a graph-based image segmentation to the spatio-temporal case.  

In short, given a video sequence, a spatio-temporal graph $ G = (V,E,W) $ is constructed, where the graph nodes $ V $ are the pixels in the video and are connected by edge $ E $ if they are within distance $r$ from each other in each frame, and $W$ measures the similarity of pixels connected by an edge.  Also, optical flow is used to add temporal motion information to the graph by connecting pixel $(x,y)$ in frame $t$ to its 9 neighbors along the backward flow $(u,v)$ in frame $t-1$. Affinities are calculated between pixels that are connected in the same frame and also between frames. The overall affinity matrix is a sparse symmetric block diagonal matrix that has the following structure: 
\begin{equation}
W =
 \begin{bmatrix}
W_{11} &  W_{12}  & \ldots & 0\\
W_{21}  &  W_{22} & \ldots & 0\\
\vdots & \vdots & \ddots & \vdots\\
0  &   0       &\ldots & W_{nn}
\end{bmatrix}
\end{equation}

Where $W_{ii}$ is the affinity matrix for frame 1, and $W_{ij}$ is the affinity matrix between frame $i$ and frame $j$. Then spectral clustering is applied to partition the affinity graph $W$ by computing its $k$ eigenvectors corresponding to the $k$ largest eigenvalues. However to make this process more efficient, only a randomly selected subset of pixels in the video is used to estimate a rank-k approximation of normalized affinity matrix $ L = D^{-\frac{1}{2}}WD^{-\frac{1}{2}}$, where $ D $ is the diagonal degree matrix defined as $D_{ii} = \sum_{j=1}^{n}W_{ij}$. Next, k-means is used to cluster the eigenvectors which gives an oversegmentation of the video in the form of 3D superpixels. Finally, a post-processing step is applied to merge the 3D superpixels and obtain the final segmentation.

In conclusion, given a RGB image $I_{k}$, the appearance-based spatio-temporal object discovery approach $\mathcal{D}(\cdot)$ generates a candidate objects set $C^{'}=C^{'}_{1},C^{'}_{2}, \cdots, C^{'}_{p}$:

\begin{equation}
\label{eq:appearance_discovery}
C^{'}=\mathcal{D}(I), I=I_{1}, I_{2} \cdots I_{K}
\end{equation}

\subsection{Robotic Manipulation}
\label{Appearance-Driven-Robotic-Manipulation}
Given a candidate objects set $C^{'}$, the system (robot and the arms) uses a motion cue to verify the validation of an object. The system randomly selects an object and then uses the robot arms to move it (grasping or poking). To interact with the object, the object pose $T_{wO}$ has to be known, which can be computed by:
\begin{equation}
T_{wO}=T_{wC} \oplus T_{CO}
\end{equation}
\noindent
where $T_{CO}$ is estimated by the relative pose between the central point of $C^{'}_{i}$ and the camera. The size of the object can be approximated by estimating the size of $C^{'}_{i}$ in $x,y,z$ direction. Given the object's pose $T_{wO}$, a grasp generator determines what grasp and pre-grasp positions are necessary to move the object. An inverse kinematics solver is used to evaluate the feasibility of the generated candidate grasps. If no valid grasp is found due to the geometry or reach-ability of the object, a poking start and end position is generated. A joint trajectory to the pre-grasp or poking start position is generated using the motion planning framework \cite{coleman2014reducing}. The trajectory is then executed by the robot. The robot then performs a straight line motion from the pre-grasp position to the grasp position, or from the poking start to poking end position. 

If grasping, the end effector constrains the object and lifts it to within proper range of the camera. A pre-defined trajectory of the arm moves (rotates \& translates) the object through the scene under the observation of the camera (head mounted). This trajectory is designed to optimize the views of the object that can be observed which allows the system to capture enough data for object learning (tracking and reconstruction).

\subsection{Motion-Based Object Verification}
\label{Motion-Based-Objects-Verification}
While the robot is manipulating the objects, the system tracks $O_{\emptyset}$ by estimating the relative pose $T_{rl}^\emptyset$ of the camera between the reference frame $I_{r}$ and the live frame $I_{l}$ via the ICP+RGB algorithm. A virtual image $I_{v}$ of $O_{\emptyset}$ can then be generated by:
\begin{equation}
\label{eq:discovery}
I_{v}=\Upsilon(\mathcal{S}_{\emptyset},T_{wl}^\emptyset)
\end{equation}
\noindent where $\Upsilon(\cdot)$ is the ray casting operator (Section \ref{sec:Reconstruction}). If any objects have relative motion against $O_{\emptyset}$, the system can discover them by extracting portions of the scene that fail to track with $O_{\emptyset}$:
\begin{equation}
I_{o}=I_{v}-I_{l}
\end{equation}

By searching the disjoint contours with a minimum number of valid pixels $d_{1}$ in $I_{o}$, a contour set $C$ of candidate objects can be obtained:

\begin{equation}
C=\Gamma(I_{v}-I_{l})=\Gamma(\Upsilon(\mathcal{S}_{\emptyset},T_{wl}^\emptyset) -I_{l}), C=C_{1},\cdots,C_{p}
\end{equation}
\noindent Here $\Gamma(\cdot)$ detects disjoint contours in an image \cite{teh1989detection}.

The motion-based objects verification pipeline allows the system to verify objects from the candidate objects set and discard negative object discovery results if the expected motion between the candidate object and the dominant object does not happen. With this process, the system generates a verified objects set $C$, which will be used for object learning.

\section{Tracking and Reconstruction}
\label{Tracking-and-Reconstruction}
For a discovered (verified) object, the proposed system tracks and reconstructs it in 2D and 3D simultaneously.

\subsection{2D Tracking}
\label{sec:2DTracking}
For $C_{j}\in C$ that matches (tracks) with $O_{j}\in O$ by the appearance, shape and motion cues \cite{ma2014unsupervised}, $C_{j}$ defines an image domain $\Omega^{j}$:

\begin{equation}
\Omega^{j}= \mathcal{G} (C_{j})
\end{equation}

\noindent Here, $\mathcal{G} (C_{j})$ inflates the image region defined by $C_{j}$. Notice that the 2D tracking process only gives a rough tracking result between $C_{j}$ and $O_{j}$ in the 2D image domain.

\subsection{2D Reconstruction.} 
\label{sec:2DReconstruction}
Given a discovered object contour $C_{i} \in C$ (matches with an object $O_{i}$), the system does not update $\mathcal{S}_{i}$ with $C_{i}$ directly. Instead, it updates the 2D SDF of $O_{i}$ by level-set-evolution first, and then uses the updated 2D SDF to track and reconstruct $O_{i}$ in 3D space.

The system avoids updating $\mathcal{S}_{i}$ with $C_{i}$ directly because: 1) the precision of the $C_{i}$ is not guaranteed. Noisy information may be introduced from the object discovery approach to $C_{i}$, which would hamper the final 3D reconstruction result if the system updates $\mathcal{S}_{i}$ with $C_{i}$ directly. 2) Multiple objects may be overlapping with each other, but it is necessary to extract the precise contours from $C_{i}$ for each object.
3) The system needs to update the object's 3D model by fusing pixels that have not yet been observed, where $C_{i}$ only matches against the known object pixels. However, LSE solves this by updating the object's 2D model based on its previous 2D model in $C_{i}$.
\begin{equation}
\label{eq:pwp-drlse-operation}
\Phi^{'}(x_{i})= \mathcal{L} (\Phi(x_{i}),\Omega^{i})
\end{equation}
Here $\mathcal{L}(\cdot)$ is the level-set-evolution (LSE) operation \cite{ma2014unsupervised} that updates the object's 2D SDF $\Phi(x_{i})$ to $\Phi(x_{i})'$ in $\Omega^{i}=\mathcal{G}(C_{i})$. The foreground (in the RGBD space) defines by $\Phi(x_{i})'$ is $\Phi^{'}_{f}(x_{i})$

\subsection{3D Tracking.} 
\label{sec:3DTracking}
Once the 2D model of an object is updated via LSE, the system tracks $\Phi^{'}_{f}(x_{i})$ against $\Phi_{f}(x_{i})$ for 3D tracking. This is done by estimating the relative camera pose $T_{rl}^i$ between $T_{wr}^i$ (pose of $\Phi_{f}(x_{i})$) and $T_{wl}^i$ (pose of $\Phi^{'}_{f}(x_{i})$) with the ICP+RGB pose estimator $\mathcal{E}(\cdot)$. This combined pose estimator $\mathcal{E}(\cdot)$ allows robust tracking for objects with either complex appearance or shape (geometry).

\begin{equation}
T^{i}_{rl} = \mathcal{E}(\Phi_{f}(x_{i}), \Phi^{'}_{f}(x_{i}))
\end{equation}

\subsection{3D Reconstruction.}
\label{sec:Reconstruction}
For an object $O_{i}$ with the updated 2D SDF $\Phi^{'}(x_{i})$, the system updates $\mathcal{S}_{i}$ by fusing every valid point $\chi=(x,y,z) \in \Phi^{'}_{f}(x_{i})$ into $\mathcal{S}_{i}$:
\begin{equation}
\mathcal{S}^{'}_{i} = \mathcal{F}(\mathcal{S}_{i}, \Phi^{'}_{f}(x_{i}), T_{wl})
\end{equation}
\noindent Here, $\mathcal{F}(\cdot)$ is the SDF Fusion operation. $T_{wl}$ is the global pose of $\Phi^{'}_{f}(x_{i})$. The system also stores intensity in $\mathcal{S}^{'}_{i}$, where it can render the zero level set surface of $\mathcal{S}$ and generate a virtual gray $I^{g}_{v}$ and depth image $I^{d}_{v}$ by ray casting $\Upsilon(\cdot)$ \cite{newcombe2011kinectfusion}:

\begin{equation}
I_{v} = \Upsilon(\mathcal{S}, T_{wc}), I_{v} = I^{g}_{v}\cup I^{d}_{v}
\end{equation}
\noindent where $T_{wc}$ is the pose of the virtual camera.

\section{System Integration}
Given input RGBD images $I$, the system discovers a candidate objects set $C^{'}$ in $I$ via the appearance-based spatio-temporal object discovery approach $\mathcal{D}(I)$:

\begin{equation}
\ C^{'}=\mathcal{D}(I), C^{'}=C^{'}_{1},C^{'}_{2},\cdots,C^{'}_{n}
\end{equation}
For $C^{'}_{i} \in C^{'}$, the system verifies it by the motion cue (via robot manipulation) and produces a verified objects set $C$.

\begin{equation}
C=\Gamma(\Upsilon(\mathcal{S}_{\emptyset},T_{wl}^\emptyset) -I_{l}), C=C_{1},\cdots,C_{n}
\end{equation}
\noindent Here, $C_{j}\in C$ defines an image domain $\Omega^{j}$ that matches with $O_{i}$:

\begin{equation}
\Omega^{j}= \mathcal{G} (C_{i})
\end{equation}
\noindent where $O_{i}$ has camera pose $T_{wr}^i$, 2D SDF $\Phi(x_{i})$ and 3D SDF $\mathcal{S}_{i}$ in the reference frame. $\Phi(x_{i})$ can be updated to $\Phi^{'}(x_{i})$ by LSE $\mathcal{L}(\cdot)$.

\begin{equation}
\Phi^{'}(x_{i})=\mathcal{L}(\Phi(x_{i}),\Omega^{j}) = \mathcal{L}(\Phi(x_{i}),\mathcal{G} (C_{i}))
\end{equation}

\noindent where the foreground defines by $\Phi^{'}(x_{i})$ is:

\begin{equation}
\Phi^{'}_{f}(x_{i}) = \mathcal{H}(\Phi^{'}(x_{i}))  = \mathcal{H}(\mathcal{L}(\Phi(x_{i}),\mathcal{G} (C_{i}))) 
\end{equation}
 
\noindent here, $\mathcal{H}(\cdot)$ extract pixels inside $\Phi^{'}(x_{i})=0$.  Now, 3D tracking is processed by estimating the relative pose $T_{rl}^i$ between $\Phi_{f}(x_{i})$ and $\Phi^{'}_{f}(x_{i})$ via the ICP+RGB pose estimator $\mathcal{E}(\cdot)$:

\begin{equation}
T_{rl}^i=\mathcal{E}(\Phi_{f}(x_{i}),\Phi^{'}_{f}(x_{i}))
\end{equation}

3D Reconstruction is then achieved by updating $\mathcal{S}_{i}$ with $\Phi^{'}_{f}(x_{i})$ by SDF Fusion $\mathcal{F}(\cdot)$:

\vspace{-4mm}
\begin{equation}
\begin{aligned}
\begin{split}
\label{eq:FinalIntegration}
\mathcal{S}'_{i}=\mathcal{F}(\mathcal{S}_{i},\Phi^{'}_{f}(x_{i}),T_{wl}^i) 
=\mathcal{F}(\mathcal{S}_{i},\mathcal{H}(\mathcal{L}(\Phi(x_{i}),\mathcal{G} (C_{i}))), \\
T_{wr}^i\oplus \mathcal{E}(\Phi_{f}(x_{i}),\mathcal{L}(\Phi(x_{i}),\mathcal{G} (C_{i}))))
\end{split}
\end{aligned}
\end{equation}

\noindent Equation \ref{eq:FinalIntegration} describes how unsupervised object discovery and learning is unified by an appearance driven ($C^{'}$), motion verification ($C$), tracking ($T_{wr}^i$) and reconstruction ($\Phi(x_{i})$ and $\mathcal{S}_{i}$) pipeline. Here, the optimization of the whole system is equivalent to the optimization of each subsystem.

\section{Results}
\begin{figure}
\centering
\includegraphics[keepaspectratio=true,scale=0.11]
{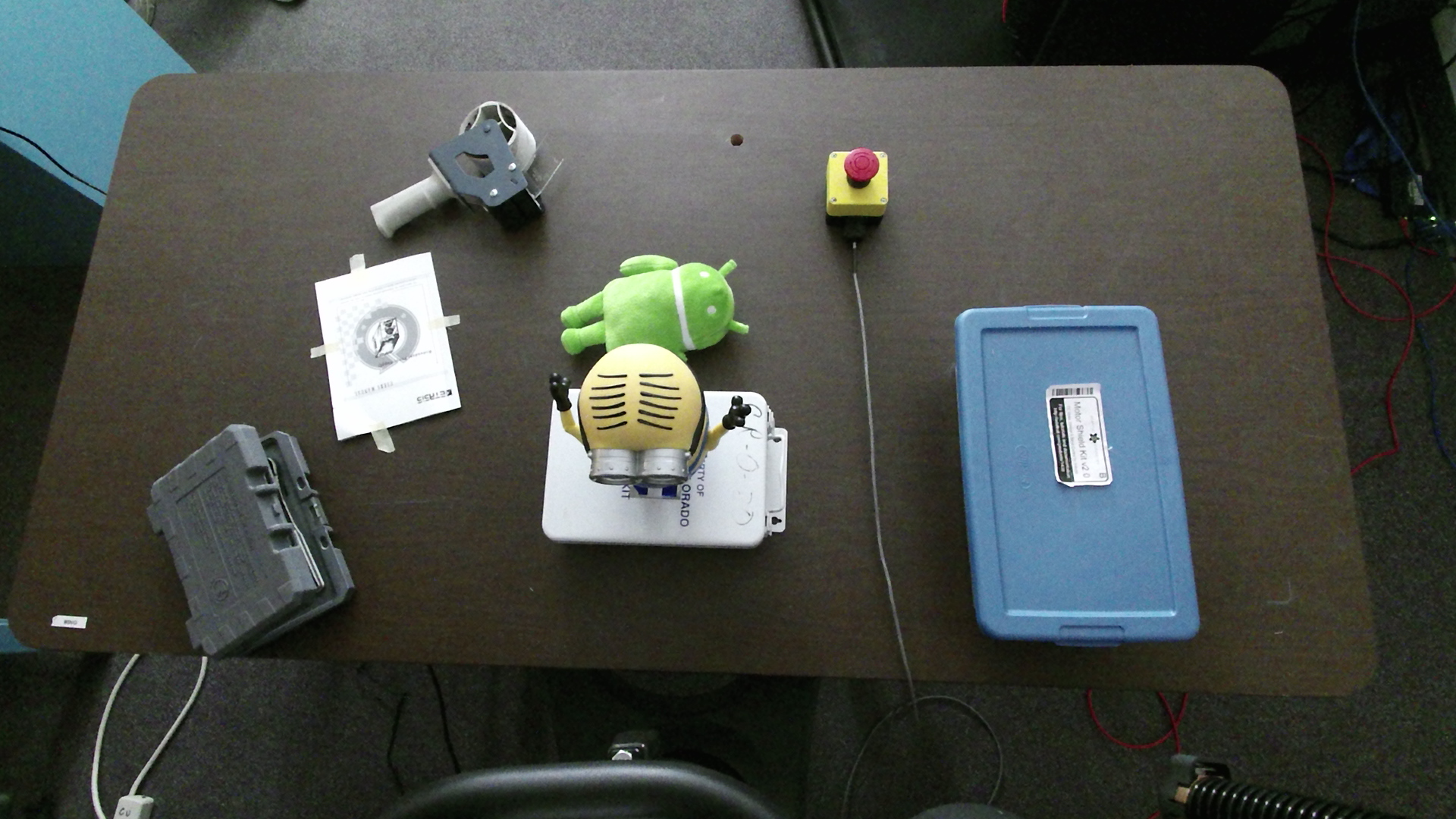}

\caption{Test scenario with objects for discovery and modeling by manipulation. The white piece of paper attached to the the desk is considered as a part of the desk (not a separate object) to test the object discovery pipeline. }
\label{fig:Scene}
\end{figure}

We evaluate the system in an unstructured scenario with different sizes, shapes and colors of objects. Notice that the white piece of paper attached to the middle-left side of the desk is considered as a part of the desk (not a separate object), which is used to test the robustness of the system. The system is tested with RGBD video streams captured at 20 FPS by a Kinect v2 camera. The camera is well calibrated and the RGB and the depth images are aligned during the experiment. Fig.\ \ref{fig:Scene} shows an example of the test scenario.

\subsection{Appearance-based object discovery}
\begin{figure*}
\centering
\subfigure[]{\includegraphics[width=.16\textwidth]{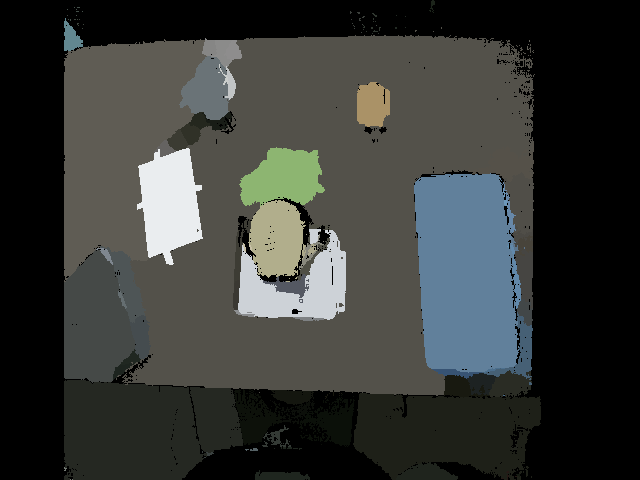}}
\subfigure[]{\includegraphics[width=.16\textwidth]{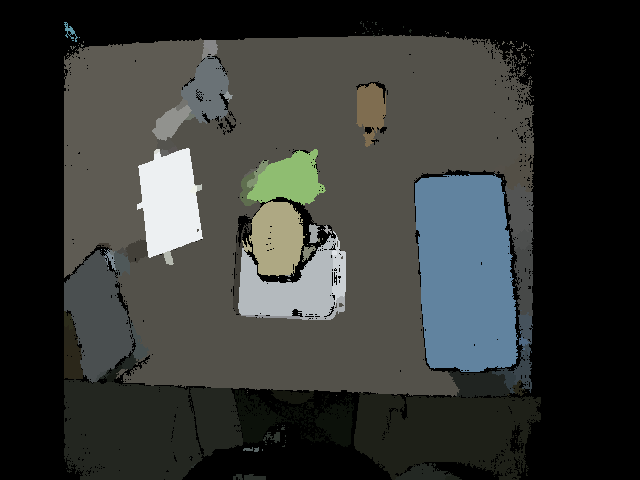}}
\subfigure[]{\includegraphics[width=.16\textwidth]{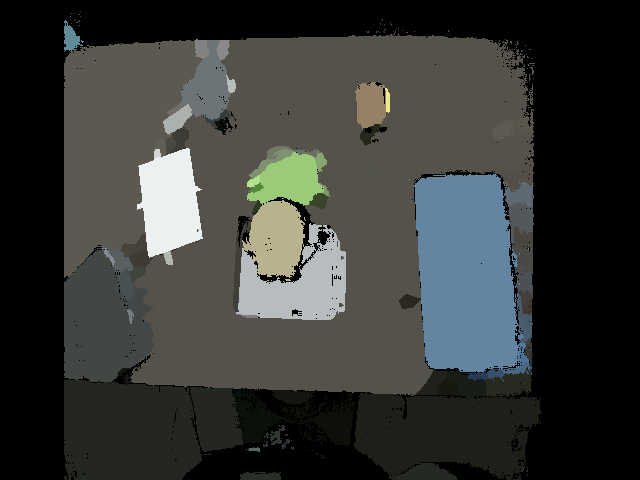}}
\subfigure[]{\includegraphics[width=.16\textwidth]{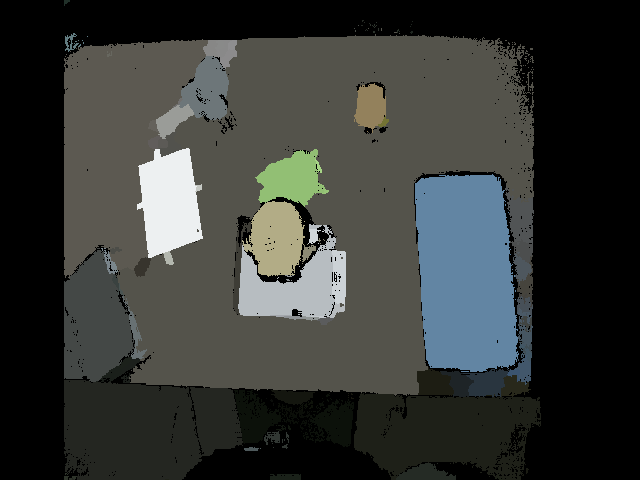}}
\subfigure[]{\includegraphics[width=.16\textwidth]{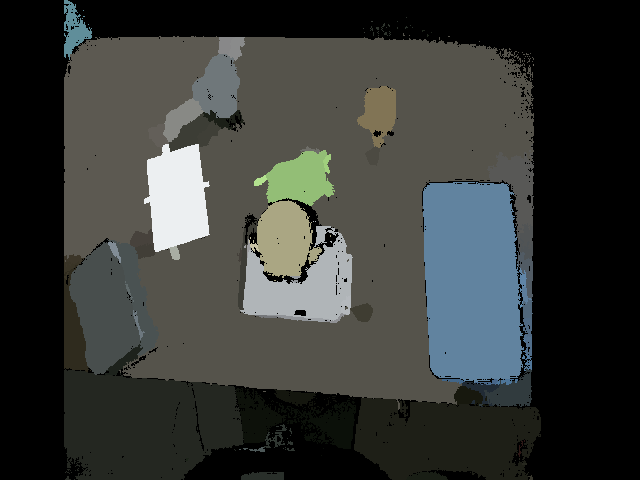}}
\subfigure[]{\includegraphics[width=.16\textwidth]{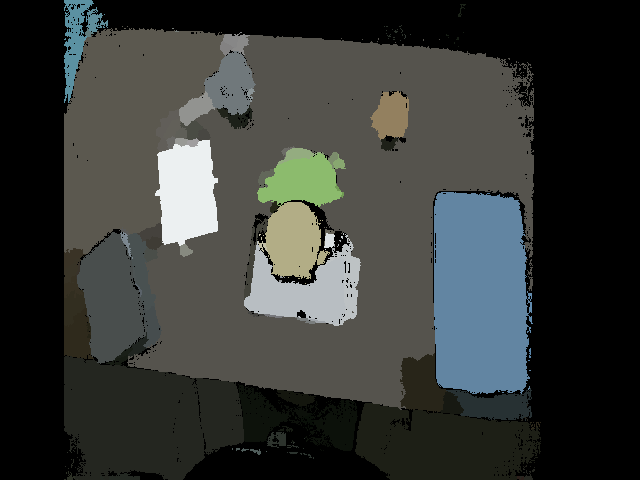}}
\\[-1pt]
\subfigure[]{\includegraphics[width=.16\textwidth]{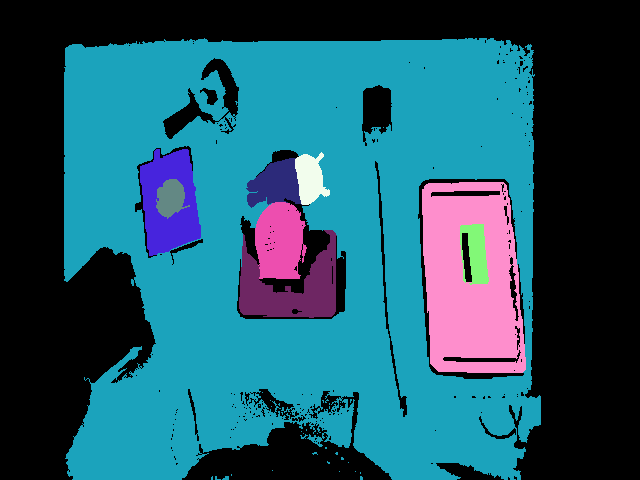}}
\subfigure[]{\label{fig:egbis-unstable-1}\includegraphics[width=.16\textwidth]{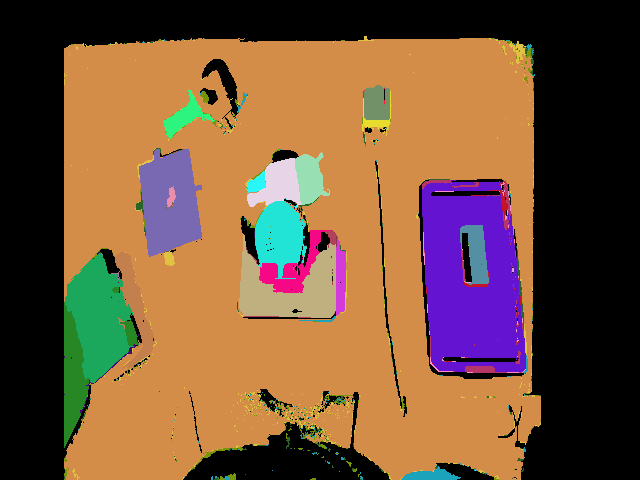}}
\subfigure[]{\includegraphics[width=.16\textwidth]{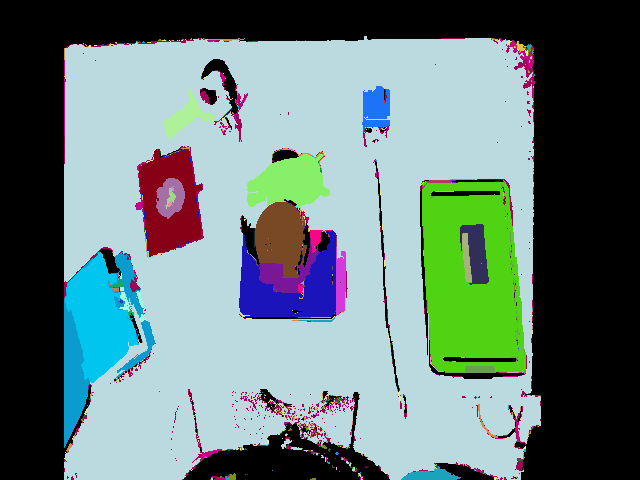}}
\subfigure[]{\includegraphics[width=.16\textwidth]{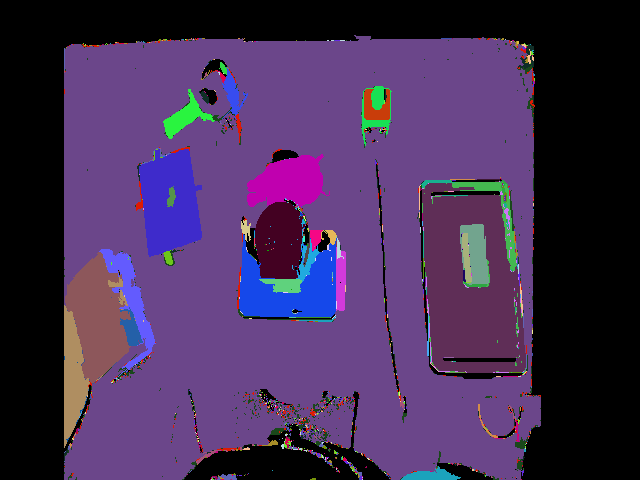}} 
\subfigure[]{\label{fig:egbisbest}\includegraphics[width=.16\textwidth]{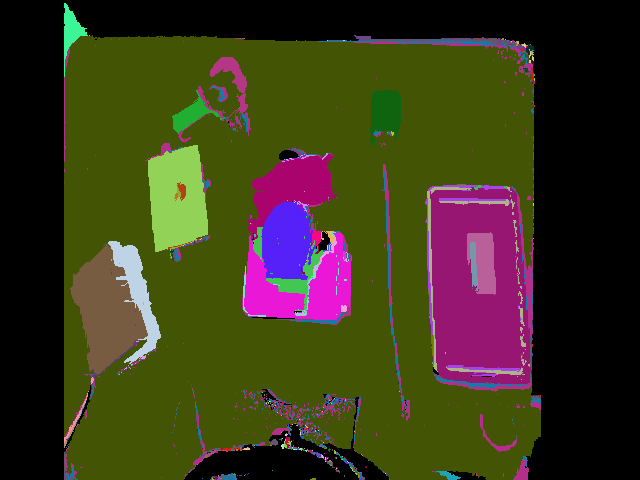}}
\subfigure[]{\label{fig:egbis-unstable-2}\includegraphics[width=.16\textwidth]{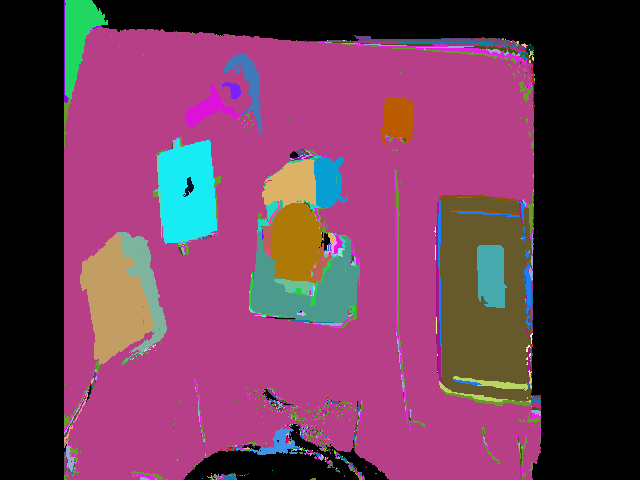}}
\\[-1pt]
\subfigure[]{\includegraphics[width=.16\textwidth]{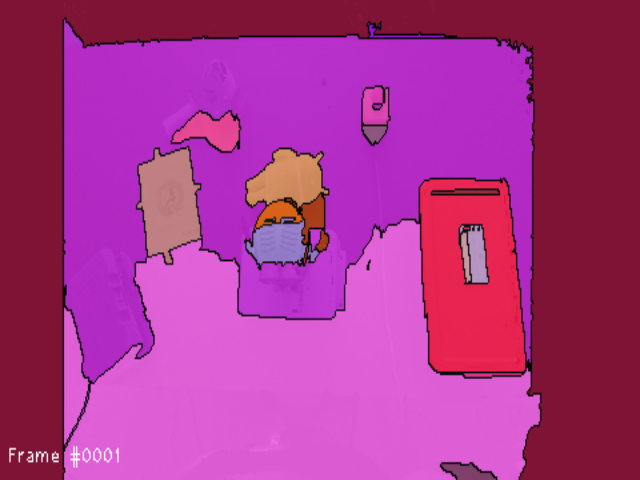}}
\subfigure[]{\includegraphics[width=.16\textwidth]{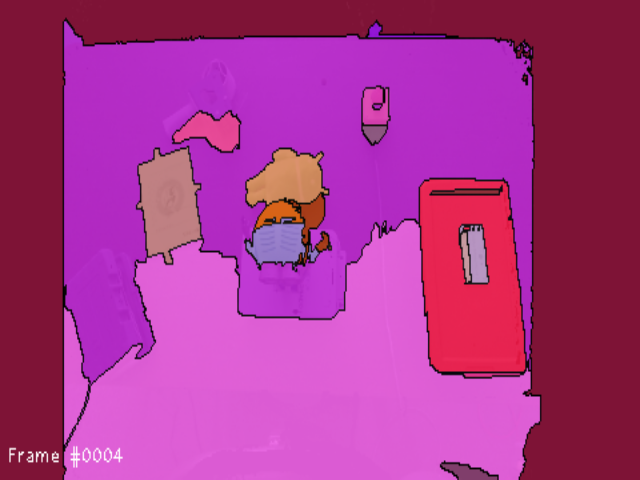}}
\subfigure[]{\includegraphics[width=.16\textwidth]{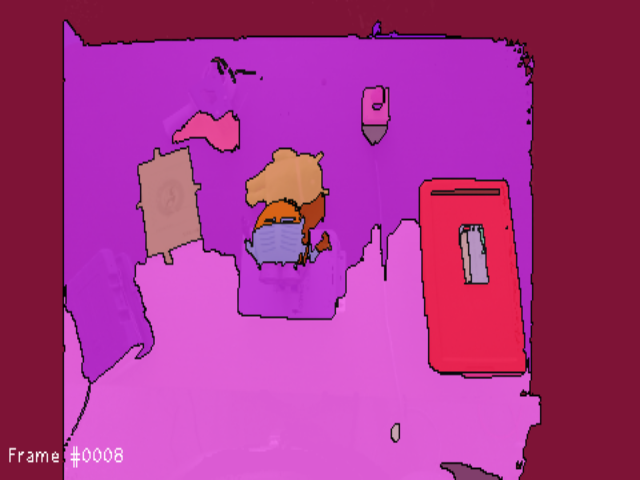}}
\subfigure[]{\includegraphics[width=.16\textwidth]{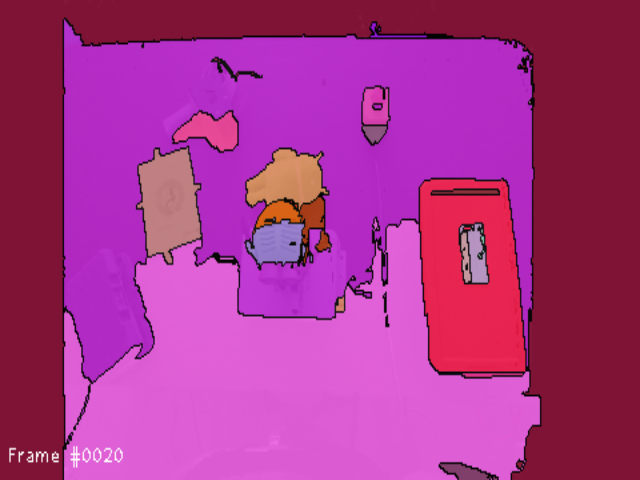}}
\subfigure[]{\includegraphics[width=.16\textwidth]{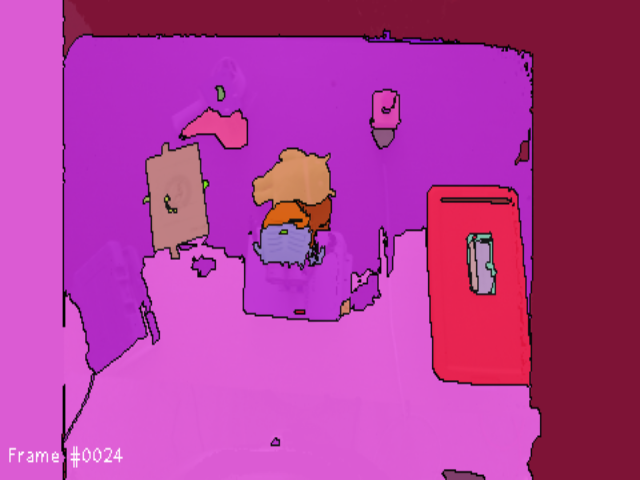}}
\subfigure[]{\includegraphics[width=.16\textwidth]{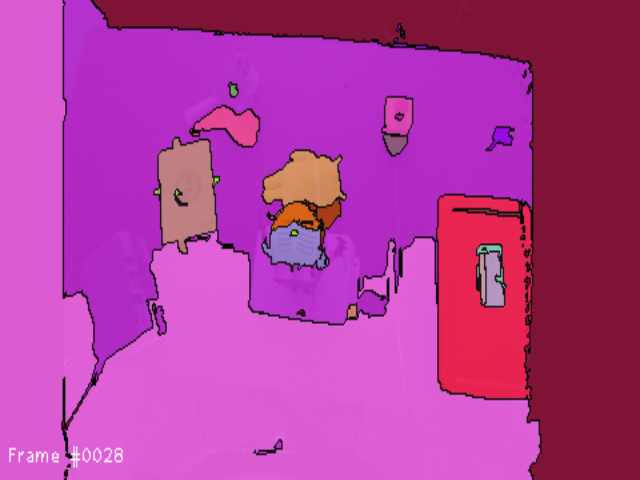}}
\\[-1pt]

\caption{Appearance-based object discovery results by the proposed method (the first row), \cite{felzenszwalb2004efficient} (the second row) and \cite{GrundmannKwatra2010} (the third row) . The proposed method has the best performance in terms of precision and recall (Fig \ref{fig:PRCurve}) of object pixels (scored using motion-verified object boundaries).}\label{fig:DiscoveryRst}
\end{figure*}

We compare our appearance-based object discovery method with two state-of-the-art object discovery approaches \cite{felzenszwalb2004efficient,GrundmannKwatra2010}. Fig. \ref{fig:DiscoveryRst} shows an example of the final discovered objects. As Fig. \ref{fig:DiscoveryRst} shows, \cite{felzenszwalb2004efficient} is able to generate good object discovery results (Fig. \ref{fig:egbisbest}) but its performance is not stable and robust (Fig. \ref{fig:egbis-unstable-1}, \ref{fig:egbis-unstable-2}). Meanwhile, \cite{GrundmannKwatra2010} generates stable object discovery results but the quality is not guaranteed (fail to discover the gray box in all sequences). The proposed method instead generates stable and precise object discovery results, which gives a good hint for the motion-based object verification pipeline.

Appearance-based object discovery methods are usually used for image segmentation, which are not designed for indoor scene object discovery. To evaluate such methods, we use a pixel level per-frame precision and recall curve approach. Here, the ground truth 3D model $\mathcal{S}_{i}$ of an object $O_{i}$ is approximated once the system learned a complete 3D model of the object, as the motion-based object discovery and learning pipeline is very robust and precise \cite{ma2014unsupervised}. In each subsequent frames, the system uses the virtual image ($D_{v}$) of $\mathcal{S}_{i}$ to approximate the ground truth label and uses $k$ different appearance-based object discovery methods to generate different discovered objects images $D^{k}_{l}$. The performance of the system is evaluated at pixel level by comparing each corresponding object's pixel $p$ in $D_{v}$ and $D^{k}_l$. The system considers a good match (true positive) if $D_{v}$ and $D_{l}$ align (match). 

\begin{figure*}
\centering
\subfigure[]{\includegraphics[width=.16\textwidth]{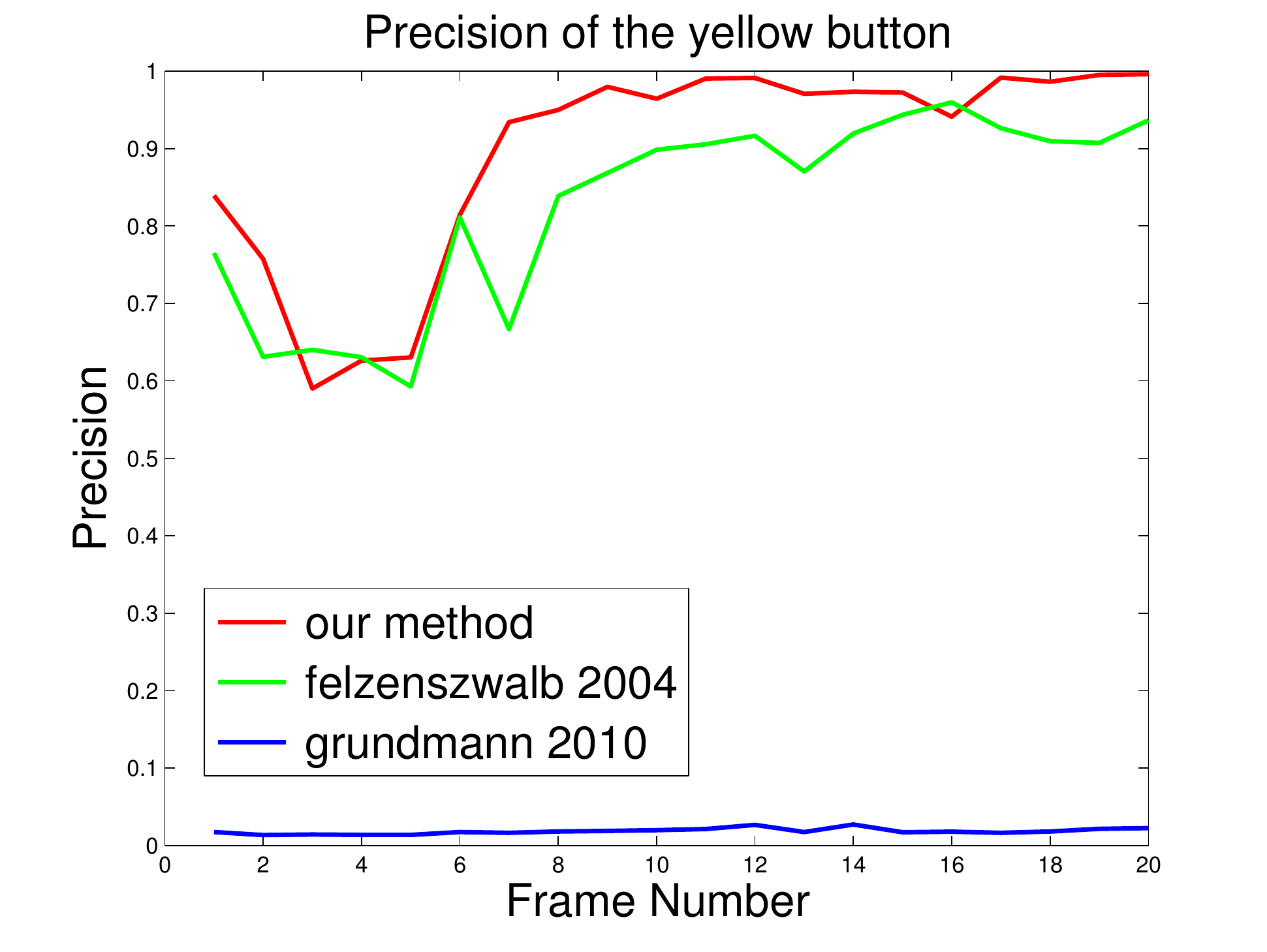}}
\subfigure[]{\includegraphics[width=.16\textwidth]{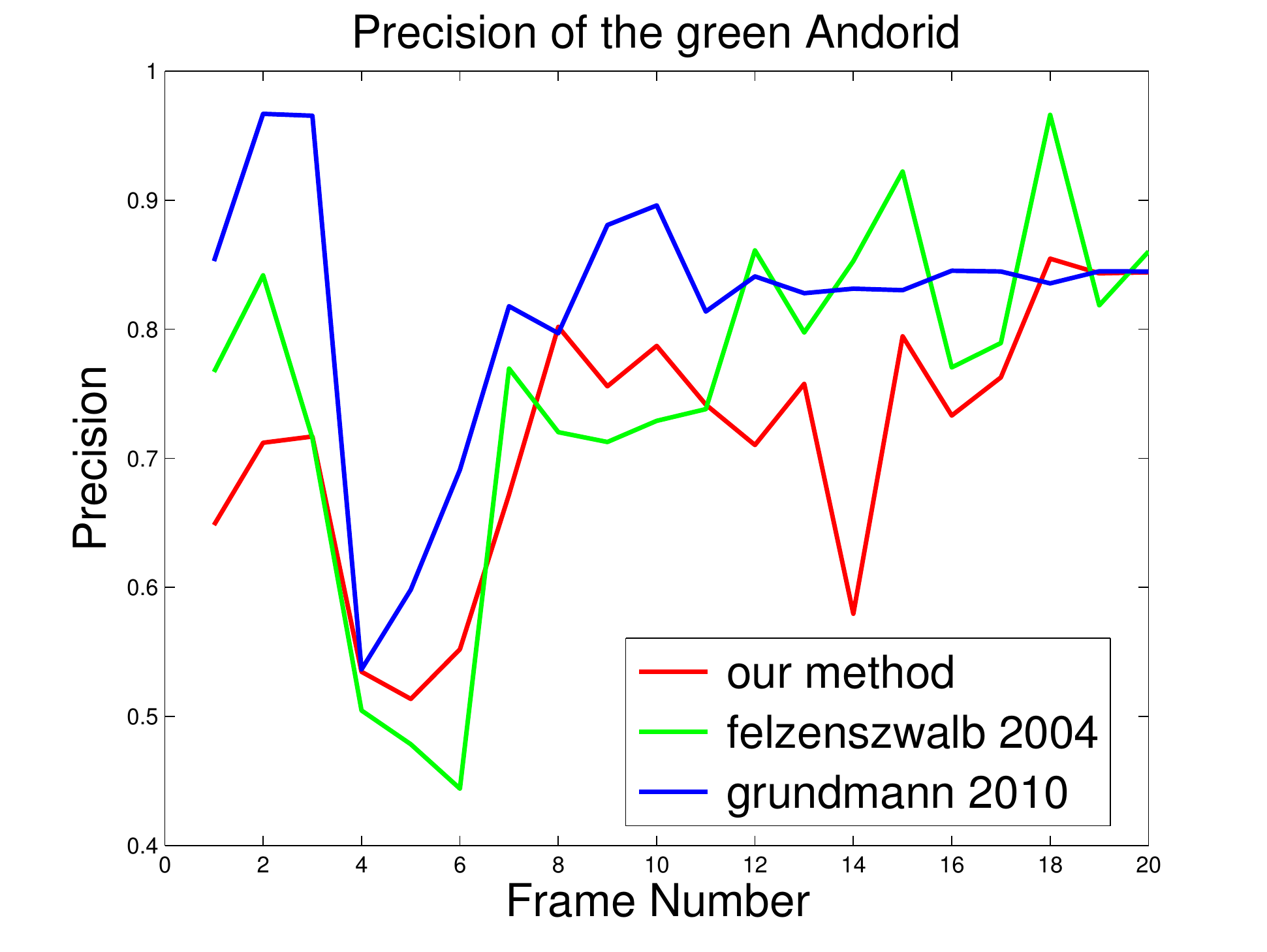}}
\subfigure[]{\includegraphics[width=.16\textwidth]{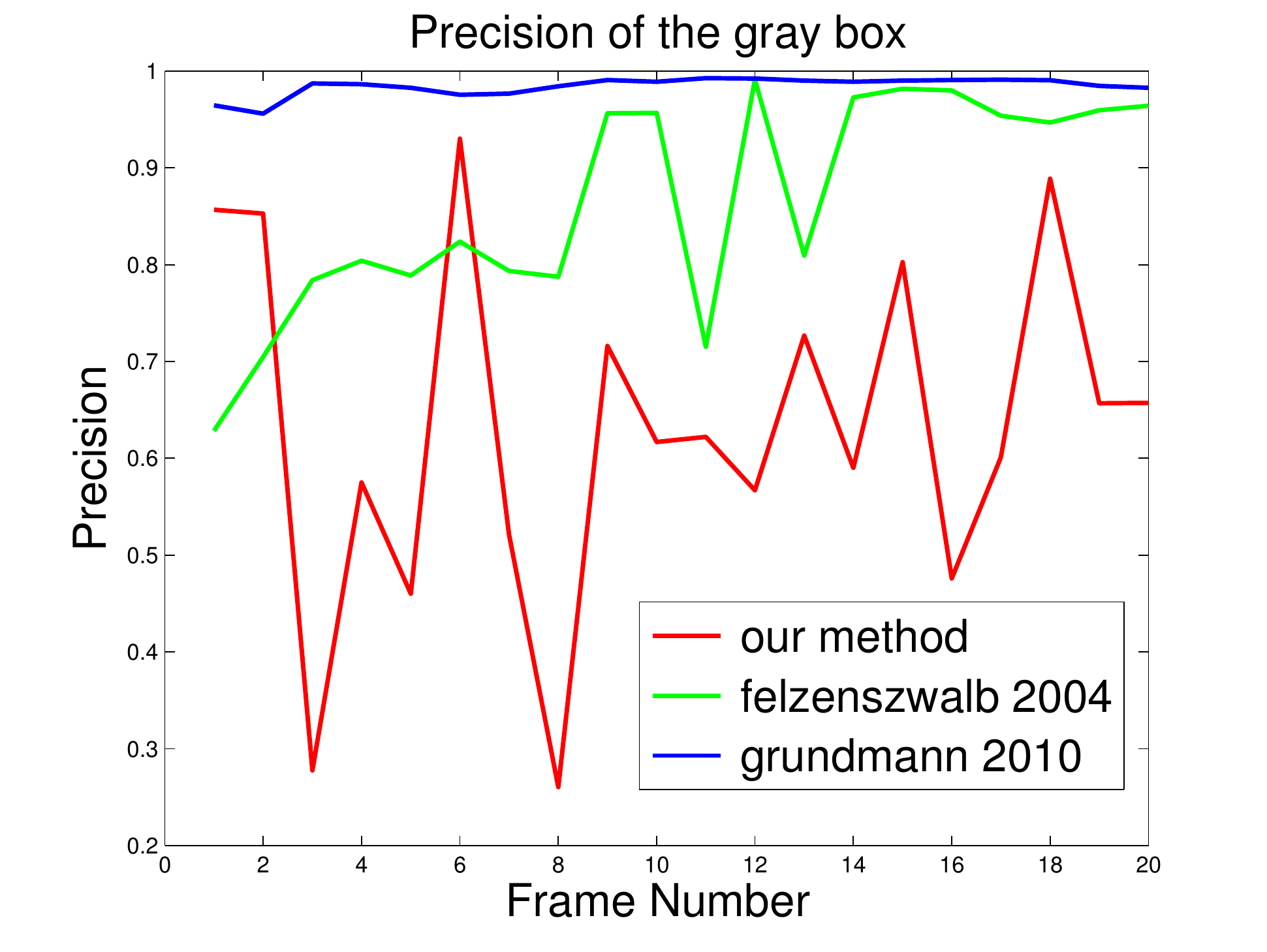}}
\subfigure[]{\includegraphics[width=.16\textwidth]{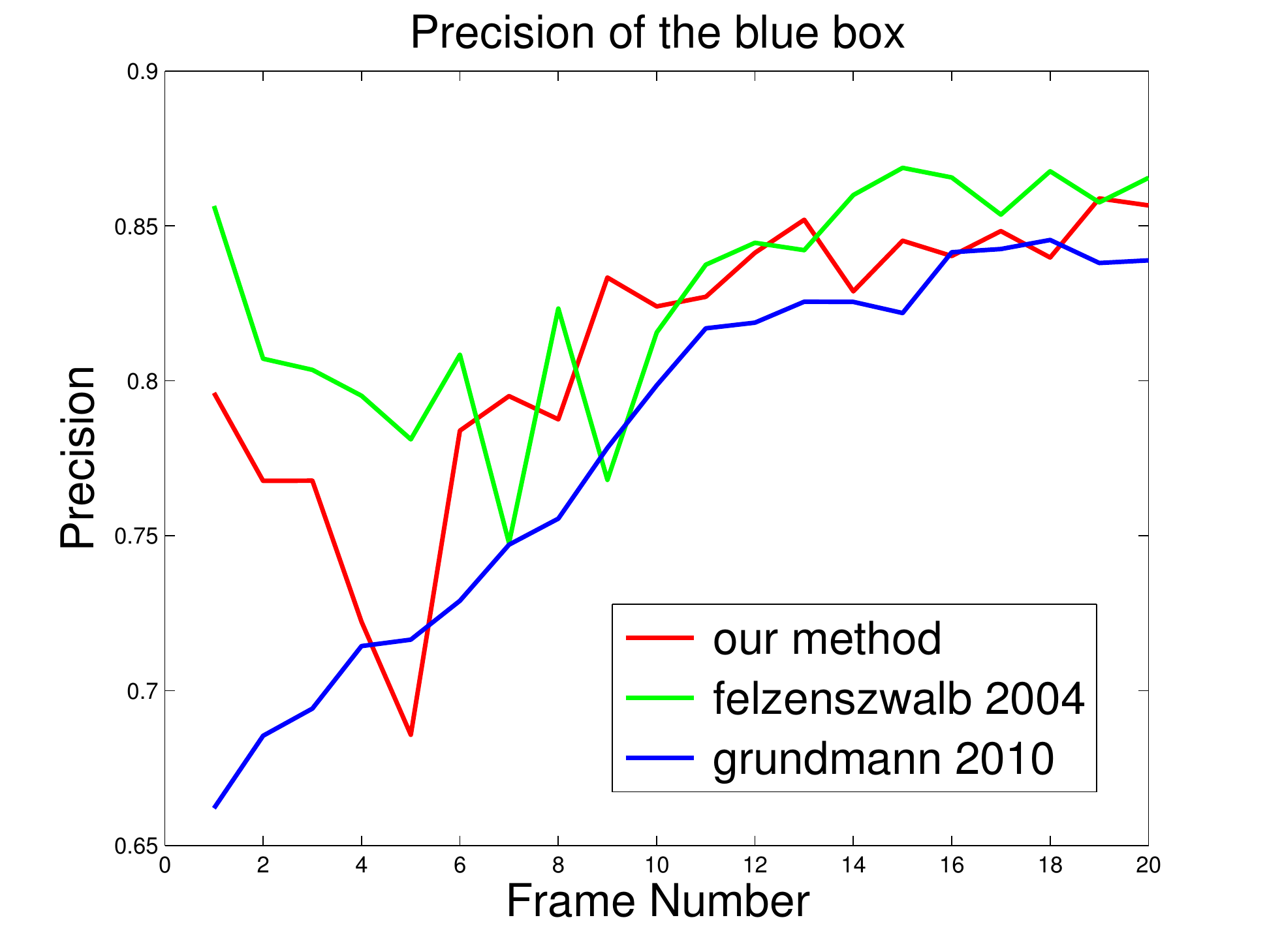}}
\subfigure[]{\includegraphics[width=.16\textwidth]{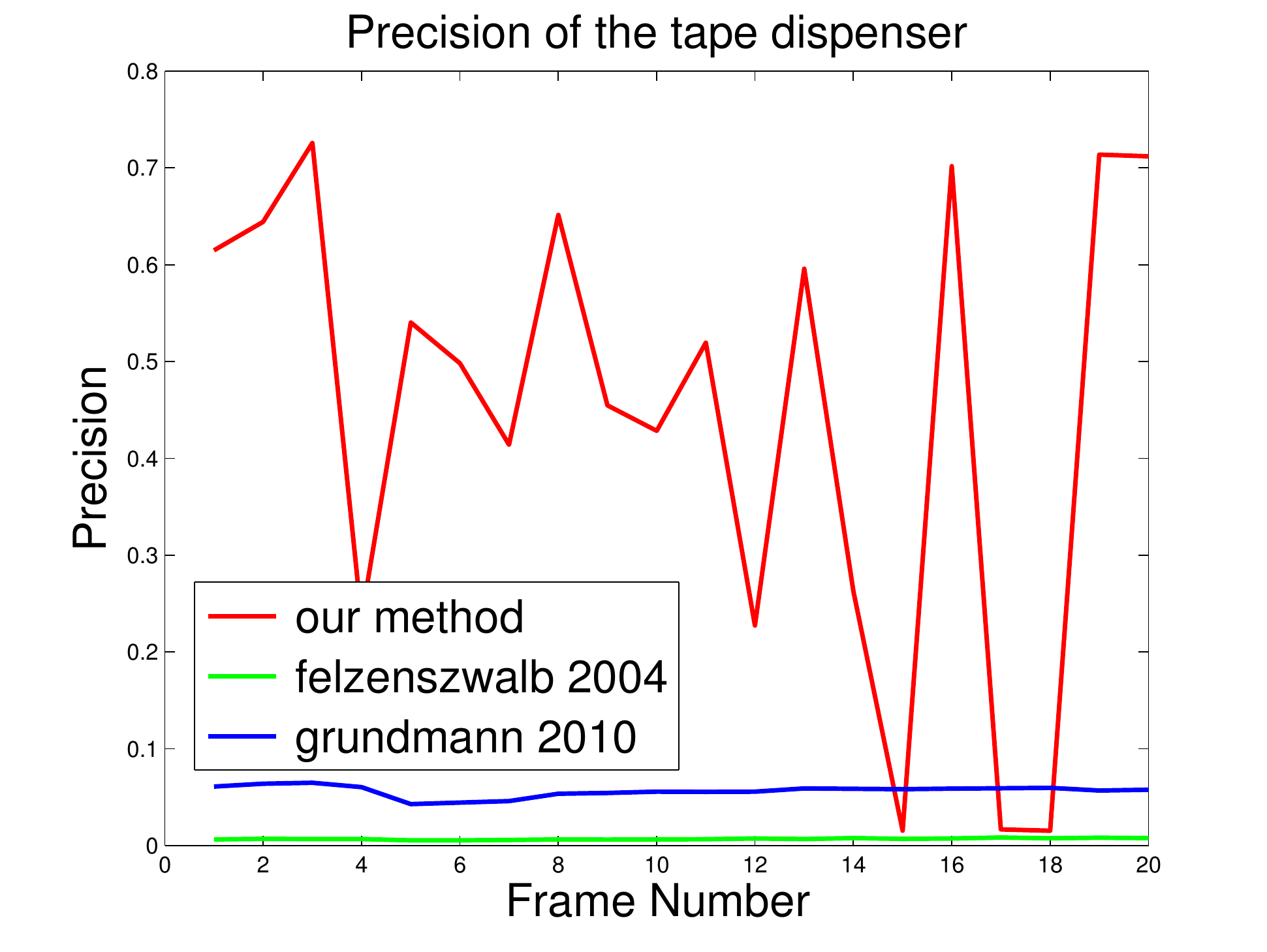}}
\subfigure[]{\includegraphics[width=.16\textwidth]{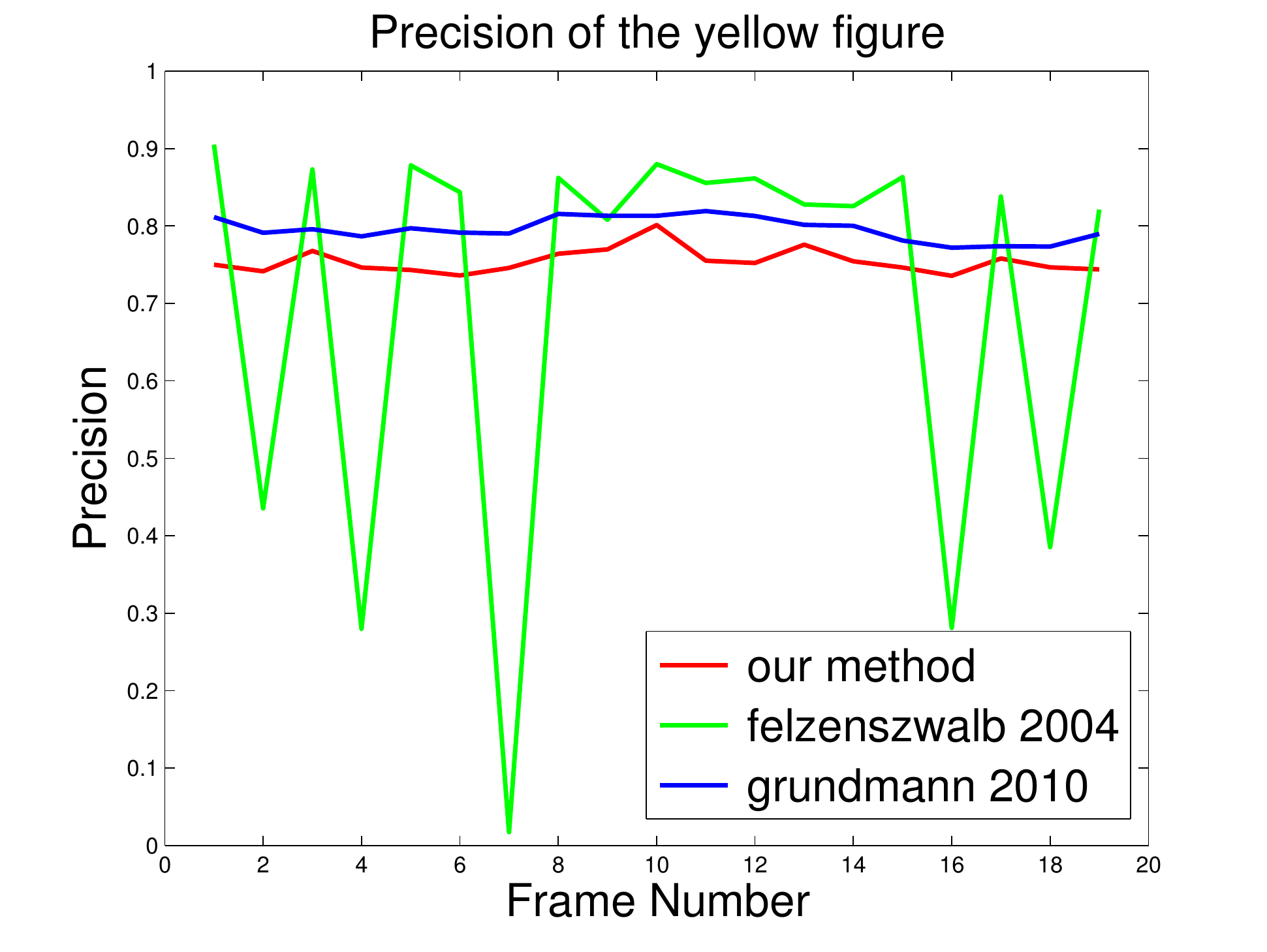}}
\\[-1pt]
\subfigure[]{\includegraphics[width=.16\textwidth]{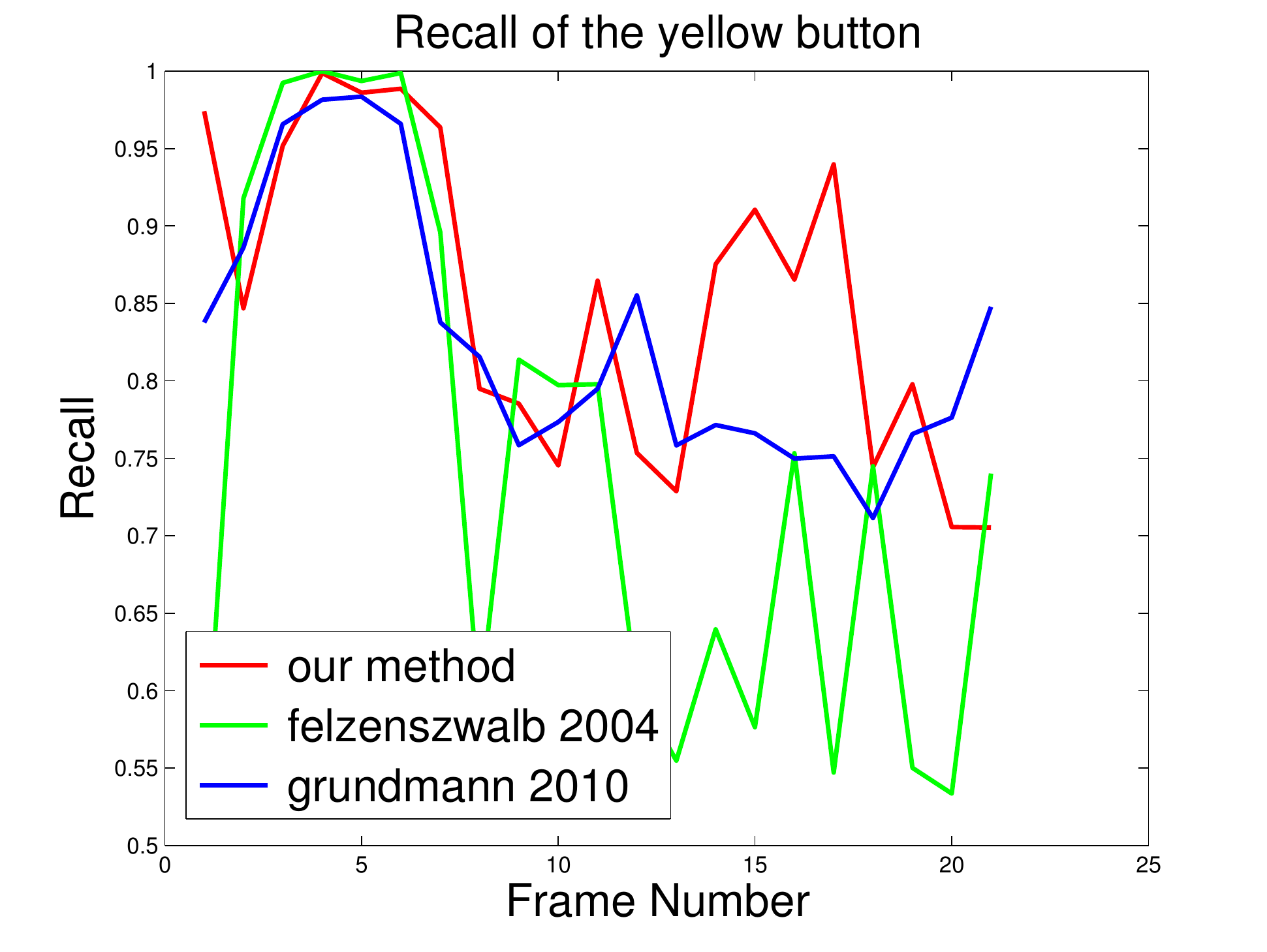}}
\subfigure[]{\includegraphics[width=.16\textwidth]{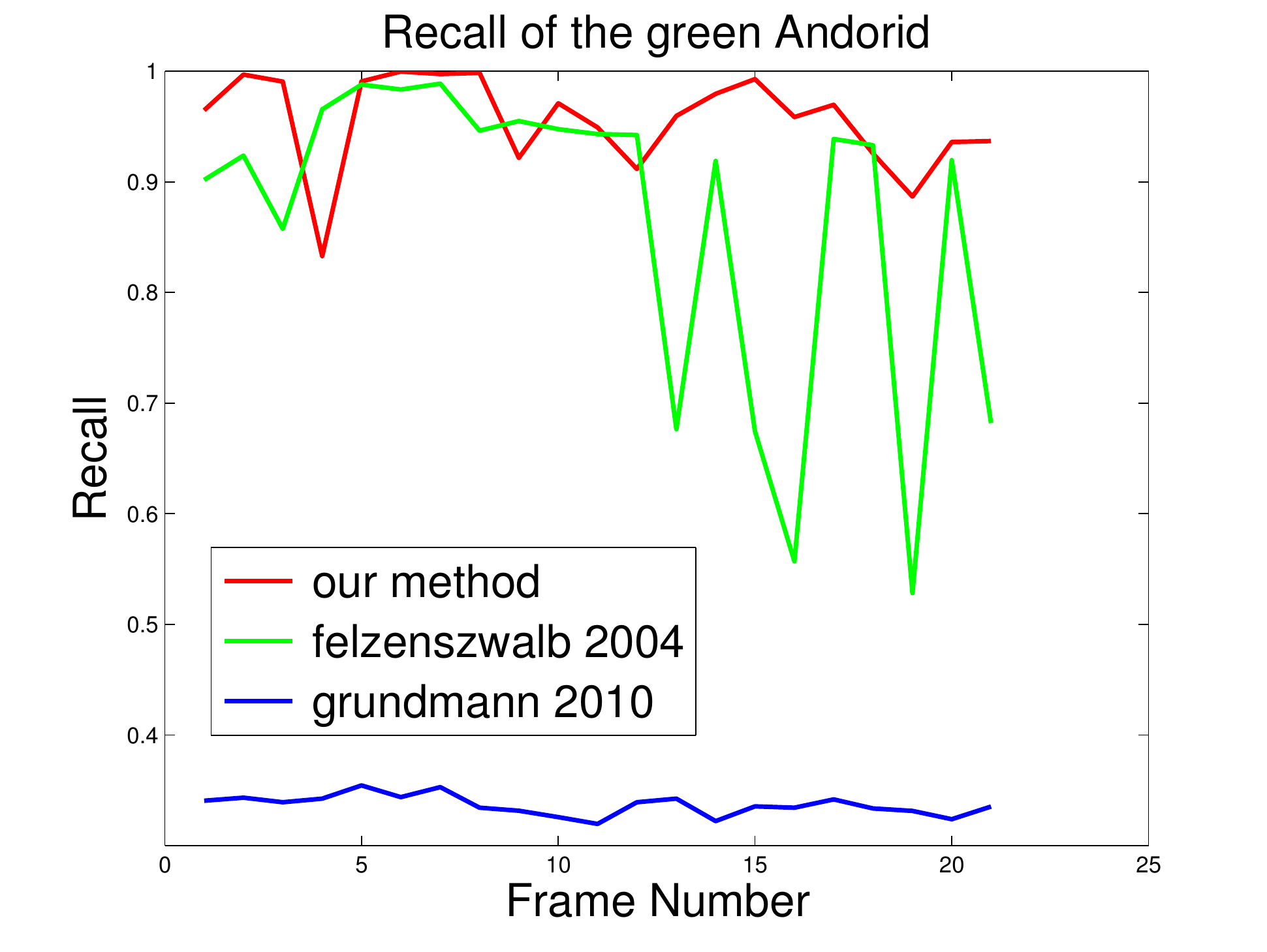}}
\subfigure[]{\includegraphics[width=.16\textwidth]{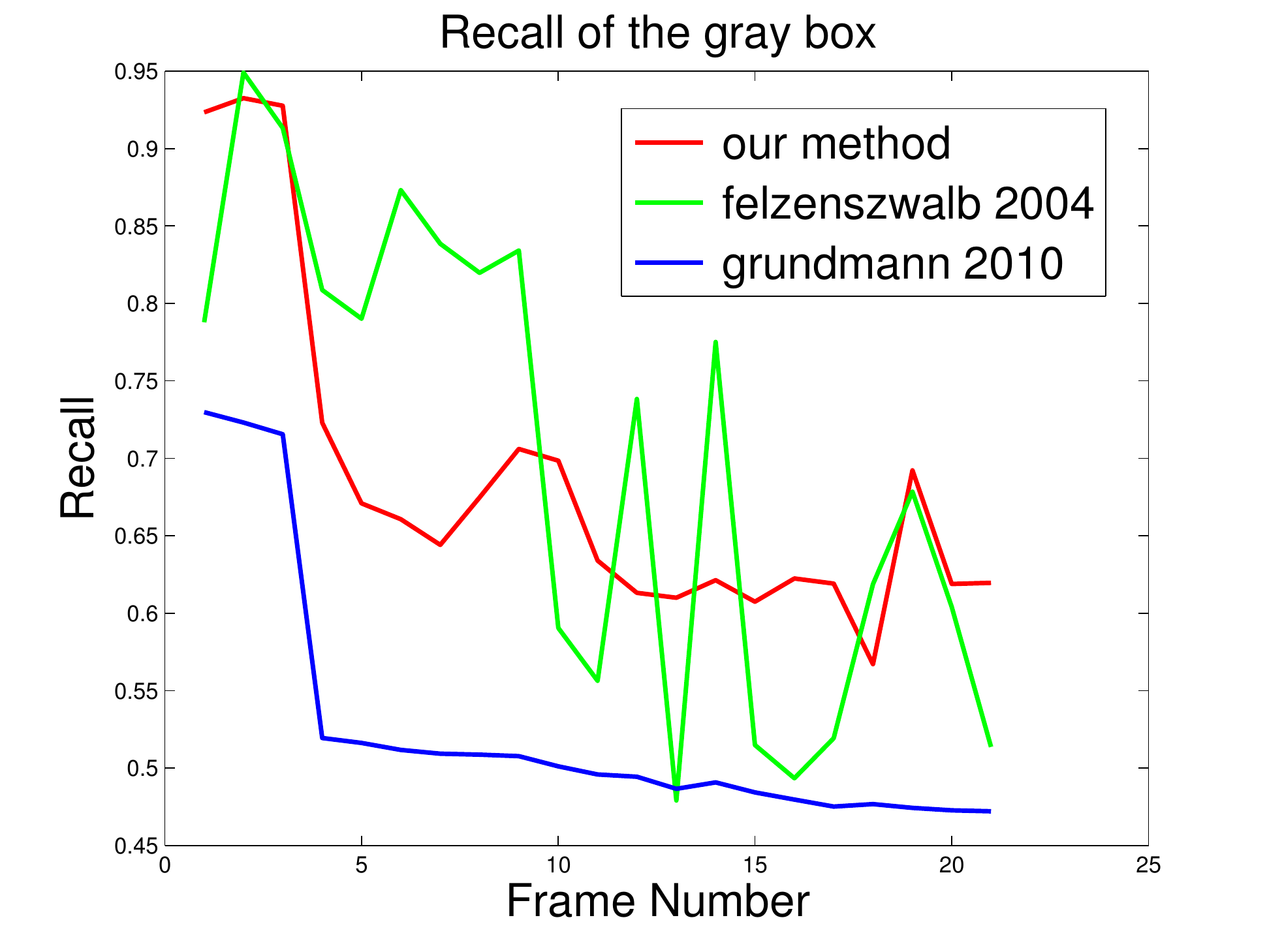}}
\subfigure[]{\includegraphics[width=.16\textwidth]{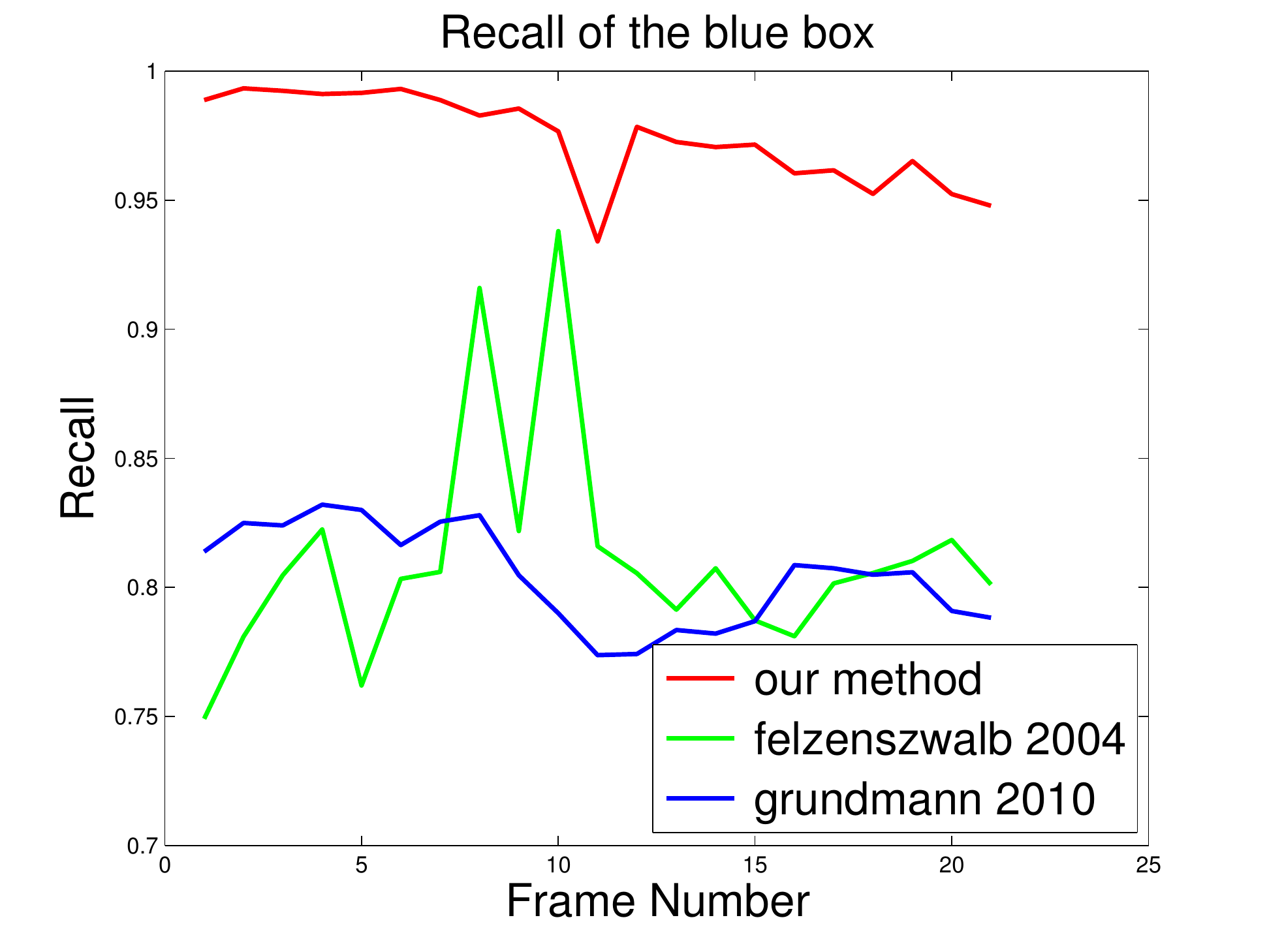}}
\subfigure[]{\includegraphics[width=.16\textwidth]{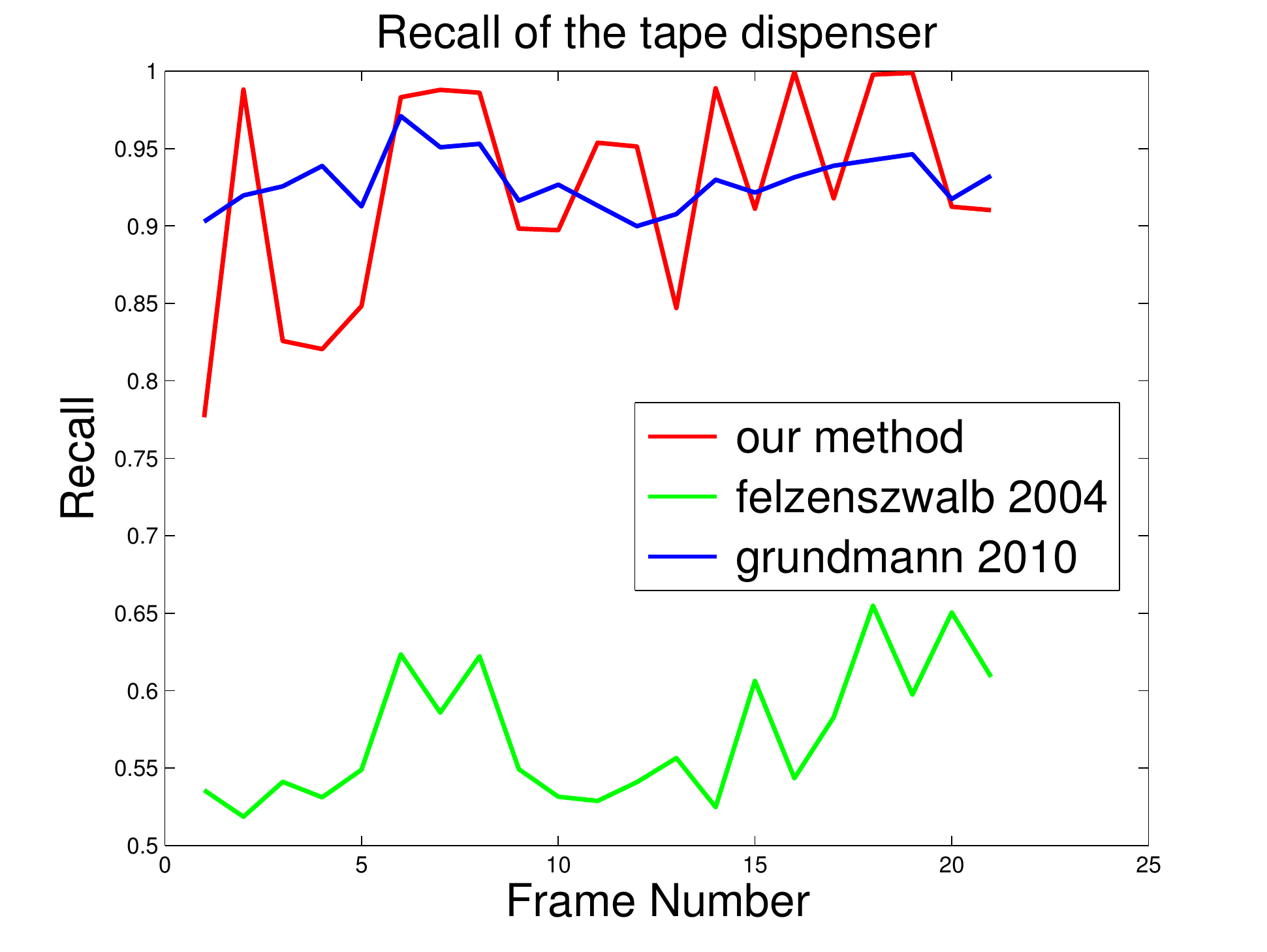}}
\subfigure[]{\includegraphics[width=.16\textwidth]{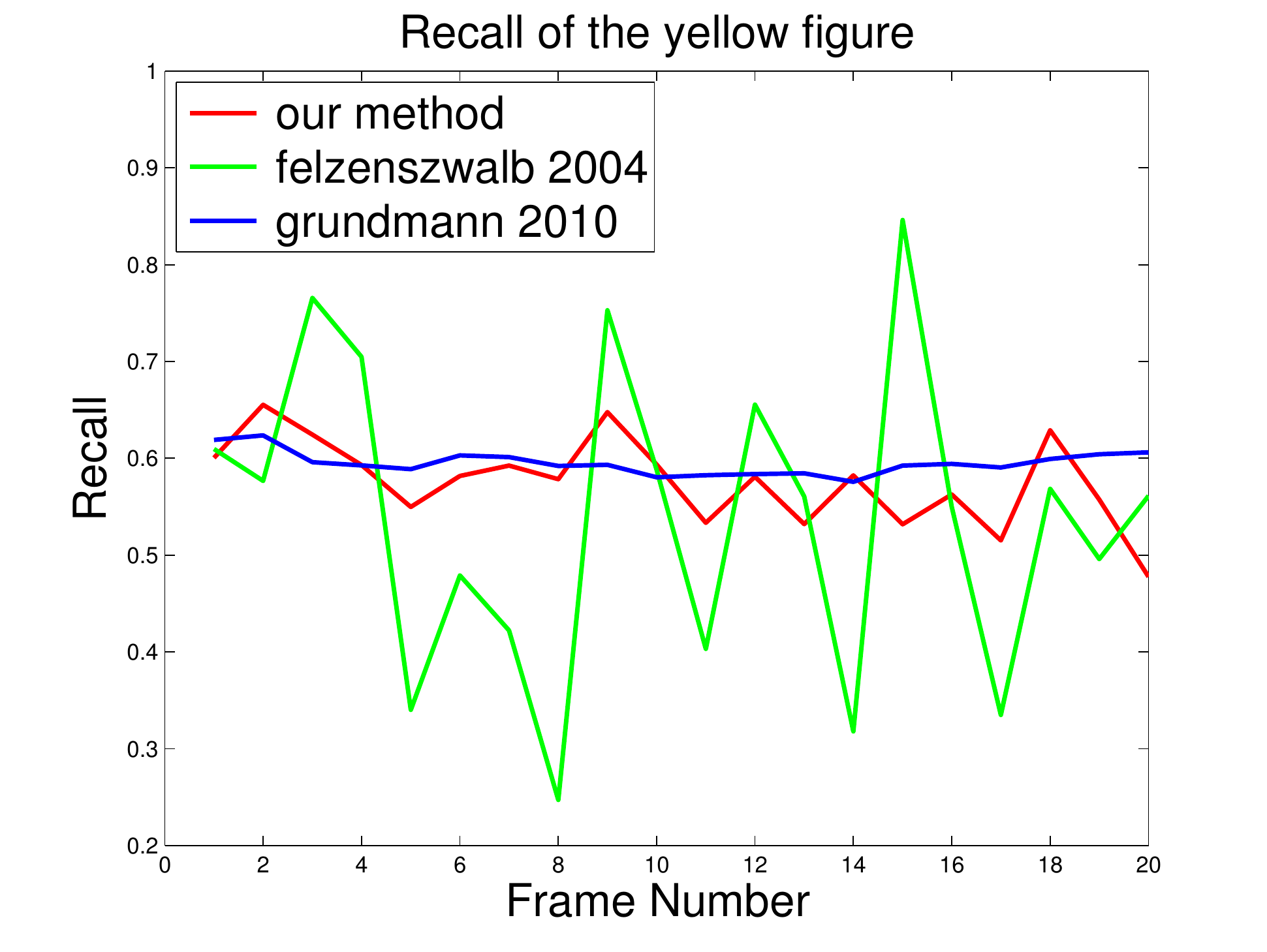}}
\\[-1pt]

\caption{Pixel level per-frame precision (the first row) and recall (the second row) curve of our approach (the red curve), \cite{felzenszwalb2004efficient} (the green curve) and \cite{GrundmannKwatra2010} (the blue curve). In general, our approach has the best performance in the test scenario.}

\label{fig:PRCurve}
\end{figure*}

We evaluate appearance-based object discovery methods with a poking and a grasping example and report the per-frame precision and recall curve (Fig. \ref{fig:PRCurve}). As Fig.\ \ref{fig:PRCurve} shows, in general our appearance-based object discovery method (the red curve) has the best performance. 

\subsection{Motion-Based Objects Verification}
\begin{figure}
\centering
\includegraphics[keepaspectratio=true,scale=0.3]{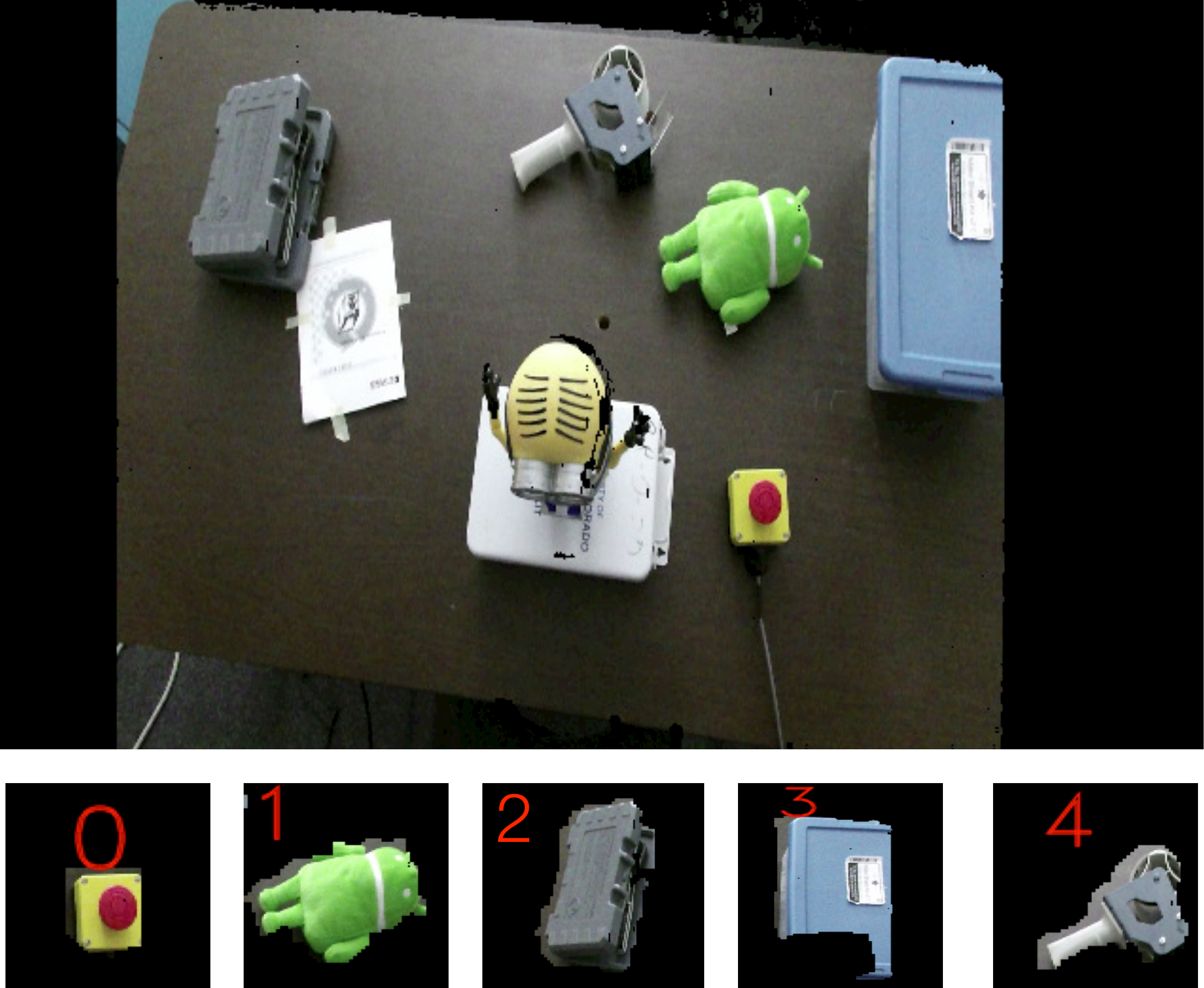}

\caption{Examples of the verified objects set (the five objects under the scene) via poking in the test scenario. The white paper from the appearance based objects discovery result (Fig. \ref{fig:DiscoveryRst}) is detected as invalid object and discarded. The yellow figure in the middle of the desk is verified and learned by grasping (Fig. \ref{fig:SingleRst}). }
\label{fig:MotionVerification}
\end{figure}

Taking the appearance-based object discovery results as a hint, the robot verifies each candidate object by motion. We test the system with a poking (the gray box, the white paper,  the tape dispenser, the yellow button, the green Android and the blue box) and a grasping (the yellow cartoon figure) experiment. Fig. \ref{fig:MotionVerification} shows an example of the final verified objects in the poking experiment. (a grasping example is available at Fig.\ \ref{fig:SingleRst}). 

Due to the limitation of the appearance-based object discovery approach, negative results (e.g. the white paper on the desk) will be produced as it considers textures (color) to be candidate objects. However, such negative discovery results can be correctly recognized by the motion-based verification pipeline.  The system will discard negative objects discovery result and only focus on the positive object discovery results. Notice that the motion-based objects verification pipeline does not discovery the blue box completely as its initial position (Fig. \ref{fig:Scene}) and final position overlaps (with the same geometry and appearance). However, the system is able recovery it via the tracking and reconstruction approaches and learn its full 3D model (Fig. \ref{fig:6-objs-learning}).

\subsection{Learning}
\begin{figure}
\centering
  \includegraphics[keepaspectratio=true,scale=0.235]{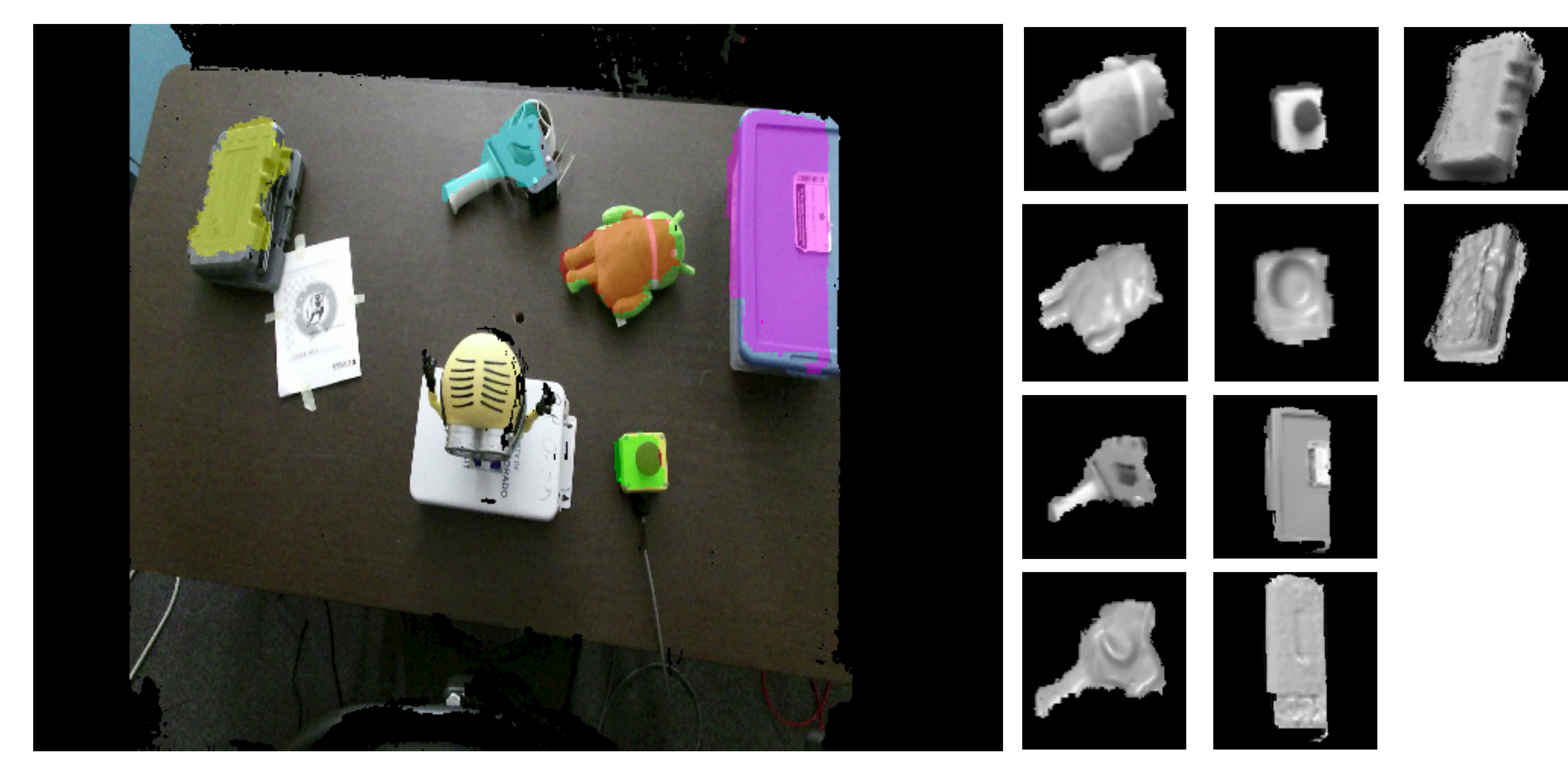}

\caption{Final learned objects of the poking experiment. The left figure shows the final tracking results (shown as different color transparent masks). The right figure shows the virtual gray and the phong shading images of the final reconstruction results.}
\label{fig:6-objs-learning}
\end{figure}

Once an object is verified, the system learns the 2D and 3D model of the object via the tracking and reconstruction pipeline. As Fig.\ \ref{fig:6-objs-learning} shows, the system is capable to learn all candidate objects verified by the poking experiment.

\begin{figure*}
\centering
\subfigure[]{\includegraphics[width=.16\textwidth]{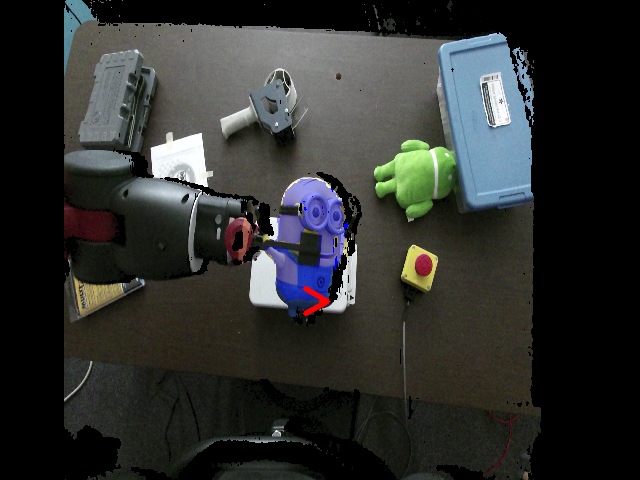}}
\subfigure[]{\includegraphics[width=.16\textwidth]{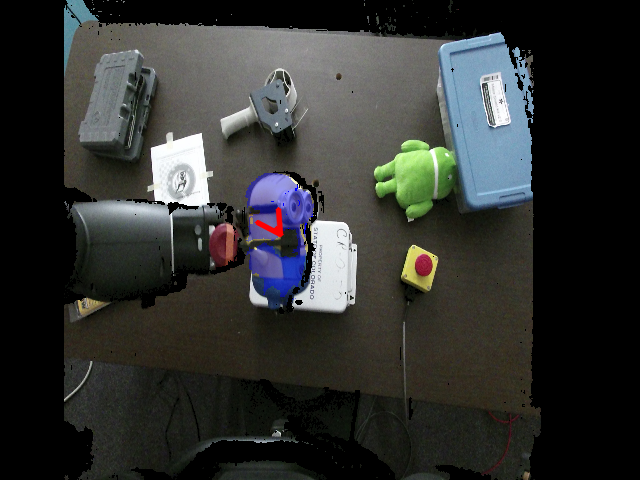}}
\subfigure[]{\includegraphics[width=.16\textwidth]{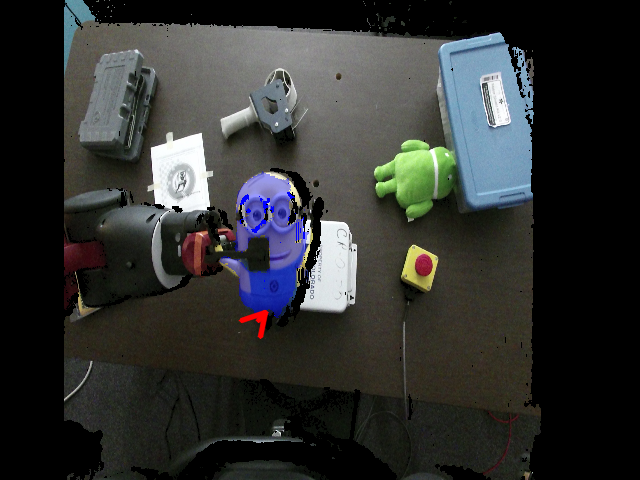}}
\subfigure[]{\includegraphics[width=.16\textwidth]{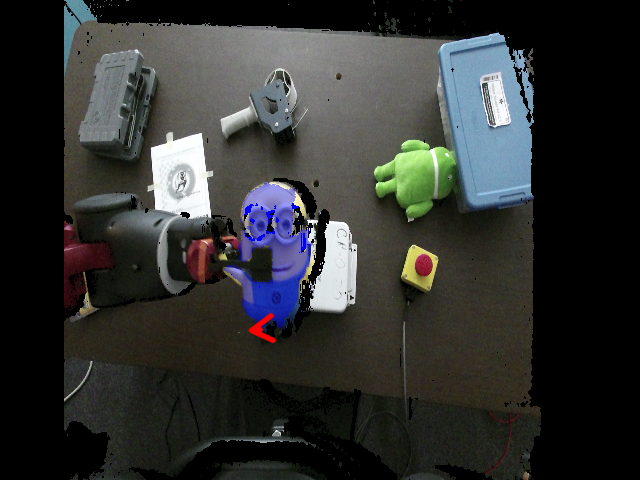}}
\subfigure[]{\includegraphics[width=.16\textwidth]{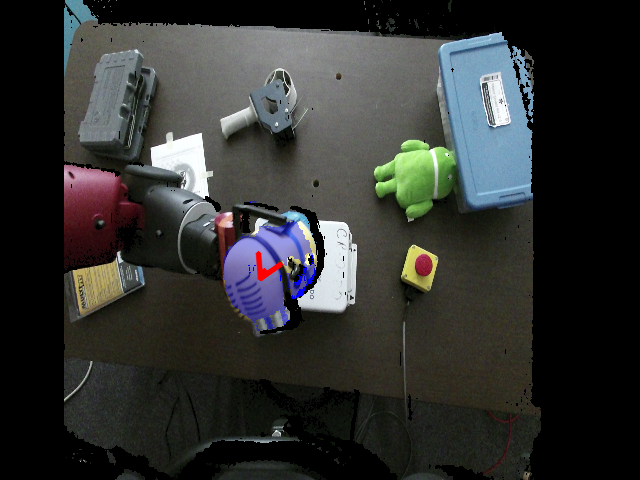}}
\subfigure[]{\includegraphics[width=.16\textwidth]{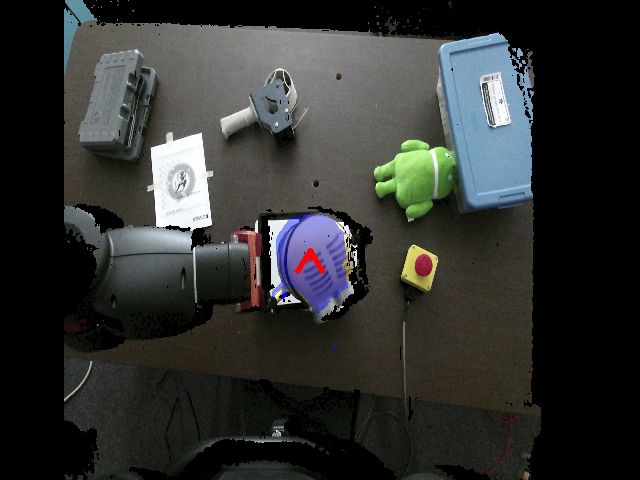}}
\\[-1pt]
\subfigure[]{\includegraphics[width=.16\textwidth]{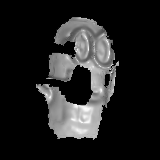}}
\subfigure[]{\includegraphics[width=.16\textwidth]{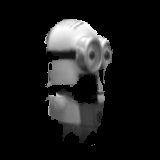}}
\subfigure[]{\includegraphics[width=.16\textwidth]{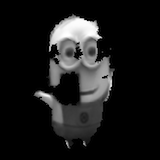}}
\subfigure[]{\includegraphics[width=.16\textwidth]{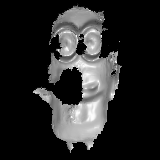}}
\subfigure[]{\includegraphics[width=.16\textwidth]{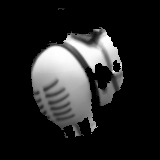}}
\subfigure[]{\includegraphics[width=.16\textwidth]{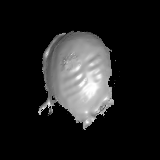}}
\\[-1pt]

\caption{Tracking (the first row, shown by blue transparent masks) and reconstruction (the second row, shown by the virtual image of the object's 3D SDF) results of a verified object (the yellow cartoon figure). In this scenario, the robot grasps an object on the desk and moves it around the scene (for 500 frames). A complete 3D model of the object is learned after the robot has enough observation of the object. The red arrow shows the object's estimated velocities.}
\label{fig:SingleRst}
\end{figure*}

To obtain a complete model of an object, the observation of multi-views of the object is required. We demonstrate this by a grasping experiment, where the robot grasps an object (the yellow cartoon figure) on the desk and moves it under the observation of the camera. A complete 3D model of the object is reconstructed when the system has a complete knowledge of the object (takes approximate 500 frames). Fig. \ref{fig:SingleRst} shows an example of the final learning result of the object via grasping.

\subsection{Time Analysis}
The system is tested using a single NVIDIA TITAN GPU, Intel i7 CPU desktop, with $640\times480$ resolution of input images. Table 1 shows our system run-time in different stages. Notice that the robotic manipulation, motion-based object verification, tracking and reconstruction approaches are processed simultaneously. The proposed system has a low running time complexity (22 Hz/object) during the learning (tracking and reconstruction) stages (GPU-based). The appearance-based object discovery approach is relatively slow, which takes approximately 10 min to discover objects using 10 frames (CPU-based). However, a similar CPU-based system \cite{grundmann2010efficient} has been implemented which runs at 1 Hz. Based on the time analysis above, it is reasonable to conjecture that our system can run in real-time after optimization.

\begin{table}[h]
\centering
\captionof{table}{System run-time} \label{tab:title2} 
\centering
    \begin{tabular}{ | l | l | l | p{5cm} |}
      \hline
      Appearance-Based Object Discovery (10 frames) & 10 min \\ \hline
      Manipulation Path Planning & 5 s \\ \hline
      Robotic Manipulation & 1 min \\ \hline
      Motion-Based Object Verification & 10 ms \\ \hline
      Tracking \& Reconstruction (one object) & 38 ms \\
      \hline
    \end{tabular}
\end{table}

\section{Failure Cases and Future Work}
Although the system is robust to many real-world operating conditions, there are several limitations of our current work. The appearance-based object discovery method depends on the difference between the foreground and the background and cannot discover objects under extremely dark/bright environments. For robot manipulation and motion-based object verification approaches, due to the limitation of the robot platform we have, the system cannot grasp or poke too small, too large, or too heavy objects. Additionally, the learning pipeline requires sufficient pixels for robust tracking, which means that sometimes the system will not learn the model for very small objects.

The success of our framework opens many areas of future work. The appearance-based object discovery pipeline can be improved by integrating object shape information. The system could use a single RGB camera by using techniques from \cite{newcombe2011dtam,newcombe2010live}. The path planning for the grasping/learning stage can be improved by a feedback system between the vision and the manipulation pipeline. The learned 3D model of objects could be shared among different robots for similar object detection by 3D tracking techniques \cite{prisacariu2012pwp3d}.

\section{Conclusion}
We present a novel unsupervised framework for a robot to discover and learn unknown objects in the scene by manipulation.  The system performs dense 3D simultaneous localization and mapping concurrently with unsupervised object discovery. 
Spatio-temporal cues and appearance cues are used to produce a set of candidate objects.  A motion-based verification strategy is used to verify candidate objects, and object appearance and shape is subsequently learned by grasping and poking.  We compare three different approaches for appearance-based object discovery and find that a novel form of spatio-temporal super-pixels gives the highest quality candidate object models in terms of precision and recall.


\balance
\bibliographystyle{IEEEtran}
\bibliography{library.bib}

\begin{thebibliography}{10}
\providecommand{\url}[1]{#1}
\csname url@rmstyle\endcsname
\providecommand{\newblock}{\relax}
\providecommand{\bibinfo}[2]{#2}
\providecommand\BIBentrySTDinterwordspacing{\spaceskip=0pt\relax}
\providecommand\BIBentryALTinterwordstretchfactor{4}
\providecommand\BIBentryALTinterwordspacing{\spaceskip=\fontdimen2\font plus
\BIBentryALTinterwordstretchfactor\fontdimen3\font minus
  \fontdimen4\font\relax}
\providecommand\BIBforeignlanguage[2]{{%
\expandafter\ifx\csname l@#1\endcsname\relax
\typeout{** WARNING: IEEEtran.bst: No hyphenation pattern has been}%
\typeout{** loaded for the language `#1'. Using the pattern for}%
\typeout{** the default language instead.}%
\else
\language=\csname l@#1\endcsname
\fi
#2}}

\bibitem{bo2013unsupervised}
L.~Bo, X.~Ren, and D.~Fox, ``Unsupervised feature learning for rgb-d based
  object recognition,'' in \emph{Experimental Robotics}.\hskip 1em plus 0.5em
  minus 0.4em\relax Springer, 2013, pp. 387--402.

\bibitem{endres2009unsupervised}
F.~Endres, C.~Plagemann, C.~Stachniss, and W.~Burgard, ``Unsupervised discovery
  of object classes from range data using latent dirichlet allocation.'' in
  \emph{Robotics: Science and Systems}, vol.~2.\hskip 1em plus 0.5em minus
  0.4em\relax Seattle, Washington;, 2009.

\bibitem{herbst2011rgb}
E.~Herbst, X.~Ren, and D.~Fox, ``Rgb-d object discovery via multi-scene
  analysis,'' in \emph{Intelligent Robots and Systems (IROS), IEEE/RSJ Int.
  Conf. on}, 2011, pp. 4850--4856.

\bibitem{meger2008curious}
D.~Meger, P.-E. Forss{\'e}n, K.~Lai, S.~Helmer, S.~McCann, T.~Southey,
  M.~Baumann, J.~J. Little, and D.~G. Lowe, ``Curious george: An attentive
  semantic robot,'' \emph{Robotics and Autonomous Systems}, vol.~56, no.~6, pp.
  503--511, 2008.

\bibitem{morisset2009leaving}
B.~Morisset, R.~B. Rusu, A.~Sundaresan, K.~Hauser, M.~Agrawal, J.-C. Latombe,
  and M.~Beetz, ``Leaving flatland: toward real-time 3d navigation,'' in
  \emph{Robotics and Automation (ICRA), IEEE Int. Conf. on}, 2009, pp.
  3786--3793.

\bibitem{lai2011large}
K.~Lai, L.~Bo, X.~Ren, and D.~Fox, ``A large-scale hierarchical multi-view
  rgb-d object dataset,'' in \emph{Robotics and Automation (ICRA), IEEE Int.
  Conf. on}, 2011, pp. 1817--1824.

\bibitem{herbst2011toward}
E.~Herbst, P.~Henry, X.~Ren, and D.~Fox, ``Toward object discovery and modeling
  via 3-d scene comparison,'' in \emph{Robotics and Automation (ICRA), 2011
  IEEE Int. Conf. on}, 2011, pp. 2623--2629.

\bibitem{collet2013exploiting}
A.~Collet, B.~Xiong, C.~Gurau, M.~Hebert, and S.~S. Srinivasa, ``Exploiting
  domain knowledge for object discovery,'' in \emph{Robotics and Automation
  (ICRA), 2013 IEEE Int. Conf. on}, 2013, pp. 2118--2125.

\bibitem{grundmann2010efficient}
M.~Grundmann, V.~Kwatra, M.~Han, and I.~Essa, ``Efficient hierarchical
  graph-based video segmentation,'' in \emph{Computer Vision and Pattern
  Recognition (CVPR), IEEE Conference on}, 2010, pp. 2141--2148.

\bibitem{ghafarianzadeh2014unsupervised}
M.~Ghafarianzadeh, M.~Blaschko, G.~Sibley, \emph{et~al.}, ``Unsupervised
  spatio-temporal segmentation with sparse spectral clustering,'' in
  \emph{British Machine Vision Conference (BMVC)}, 2014.

\bibitem{ma2014unsupervised}
L.~Ma and G.~Sibley, ``Unsupervised dense object discovery, detection, tracking
  and reconstruction,'' in \emph{Computer Vision--ECCV 2014}, 2014, pp. 80--95.

\bibitem{mason2012object}
J.~Mason, B.~Marthi, and R.~Parr, ``Object disappearance for object
  discovery,'' in \emph{Intelligent Robots and Systems (IROS), IEEE/RSJ Int.
  Conf. on}, 2012, pp. 2836--2843.

\bibitem{renstar3d}
C.~Y. Ren, V.~Prisacariu, D.~Murray, and I.~Reid, ``{STAR3D}: Simultaneous
  tracking and reconstruction of 3{D} objects using {RGB-D} data.''

\bibitem{dame2013dense}
A.~Dame, V.~A. Prisacariu, C.~Y. Ren, and I.~Reid, ``Dense reconstruction using
  3d object shape priors,'' in \emph{Computer Vision and Pattern Recognition
  (CVPR), 2013 IEEE Conference on}.\hskip 1em plus 0.5em minus 0.4em\relax
  IEEE, 2013, pp. 1288--1295.

\bibitem{bao2013dense}
S.~Y. Bao, M.~Chandraker, Y.~Lin, and S.~Savarese, ``Dense object
  reconstruction with semantic priors,'' in \emph{Computer Vision and Pattern
  Recognition (CVPR), IEEE Conference on}, 2013, pp. 1264--1271.

\bibitem{triebel2010segmentation}
R.~Triebel, J.~Shin, R.~Siegwart, R.~Y. Siegwart, and R.~Y. Siegwart,
  ``Segmentation and unsupervised part-based discovery of repetitive objects,''
  in \emph{Robotics: Science and Systems}, vol.~2, 2010.

\bibitem{modayil2004bootstrap}
J.~Modayil and B.~Kuipers, ``Bootstrap learning for object discovery,'' in
  \emph{Intelligent Robots and Systems, 2004.(IROS 2004). Proceedings. 2004
  IEEE/RSJ Int. Conf. on}, vol.~1, 2004, pp. 742--747.

\bibitem{coleman2014reducing}
D.~Coleman, I.~Sucan, S.~Chitta, and N.~Correll, ``Reducing the barrier to
  entry of complex robotic software: a moveit! case study,'' \emph{Journal of
  Software Engineering for Robotics}, 2014.

\bibitem{teh1989detection}
C.-H. Teh and R.~T. Chin, ``On the detection of dominant points on digital
  curves,'' \emph{Pattern Analysis and Machine Intelligence, IEEE Transactions
  on}, vol.~11, no.~8, pp. 859--872, 1989.

\bibitem{newcombe2011kinectfusion}
R.~A. Newcombe, A.~J. Davison, S.~Izadi, P.~Kohli, O.~Hilliges, J.~Shotton,
  D.~Molyneaux, S.~Hodges, D.~Kim, and A.~Fitzgibbon, ``Kinectfusion: Real-time
  dense surface mapping and tracking,'' in \emph{Mixed and augmented reality
  (ISMAR), 10th IEEE Int. Symp. on}, 2011, pp. 127--136.

\bibitem{felzenszwalb2004efficient}
P.~F. Felzenszwalb and D.~P. Huttenlocher, ``Efficient graph-based image
  segmentation,'' \emph{International Journal of Computer Vision}, vol.~59,
  no.~2, pp. 167--181, 2004.

\bibitem{GrundmannKwatra2010}
M.~Grundmann, V.~Kwatra, M.~Han, and I.~Essa, ``Efficient hierarchical graph
  based video segmentation,'' \emph{IEEE CVPR}, 2010.

\bibitem{newcombe2011dtam}
R.~A. Newcombe, S.~J. Lovegrove, and A.~J. Davison, ``Dtam: Dense tracking and
  mapping in real-time,'' in \emph{Computer Vision (ICCV), IEEE Int. Conf. on},
  2011, pp. 2320--2327.

\bibitem{newcombe2010live}
R.~A. Newcombe and A.~J. Davison, ``Live dense reconstruction with a single
  moving camera,'' in \emph{Computer Vision and Pattern Recognition (CVPR),
  2010 IEEE Conference on}, 2010, pp. 1498--1505.

\bibitem{prisacariu2012pwp3d}
V.~A. Prisacariu and I.~D. Reid, ``Pwp3d: Real-time segmentation and tracking
  of 3d objects,'' \emph{International journal of computer vision}, vol.~98,
  no.~3, pp. 335--354, 2012.

\end{thebibliography}

\end{document}